%% file: PaperForReview.tex
\documentclass[10pt,twocolumn,letterpaper]{article}

\usepackage{wacv}              

\usepackage{graphicx}
\usepackage{amsmath}
\usepackage{amssymb}
\usepackage{booktabs}
\usepackage{times}
\usepackage{epsfig}
\usepackage{algorithm} 
\usepackage{algorithmicx}
\usepackage{xcolor}
\usepackage{enumitem}
\usepackage[noend]{algpseudocode}
\algdef{SE}[DOWHILE]{Do}{doWhile}{\algorithmicdo}[1]{\algorithmicwhile\ #1}

\usepackage[pagebackref,breaklinks,colorlinks]{hyperref}

\usepackage[capitalize]{cleveref}
\crefname{section}{Sec.}{Secs.}
\Crefname{section}{Section}{Sections}
\Crefname{table}{Table}{Tables}
\crefname{table}{Tab.}{Tabs.}

\usepackage{tabu}

\def\wacvPaperID{818} 
\def\confName{WACV}
\def\confYear{2024}

\begin{document}

\title{Boosting Weakly Supervised Object Detection using Fusion and Priors from Hallucinated Depth}

\newcommand*{\affaddr}[1]{#1} 
\newcommand*{\affmark}[1][*]{\textsuperscript{#1}}

\author{%
Cagri Gungor\affmark[1] and Adriana Kovashka\affmark[1,2]\\
\affaddr{\affmark[1]Intelligent Systems Program}, \affaddr{\affmark[2]Department of Computer Science}\\
\affaddr{University of Pittsburgh}\\
{\tt\small cagri.gungor@pitt.edu}, {\tt\small kovashka@cs.pitt.edu}\\
\affaddr{ \small \url{https://cagrigungor.github.io/WSOD-AMPLIFIER/}}%
}
\maketitle

\begin{abstract}
   Despite recent attention to depth for various tasks, it is still an unexplored modality for weakly-supervised object detection (WSOD). We propose an amplifier method for enhancing the performance of WSOD by integrating depth information. Our approach can be applied to different WSOD methods based on multiple-instance learning, without necessitating additional annotations or inducing large computational cost. Our proposed method employs monocular depth estimation to obtain hallucinated depth information, which is then incorporated into a Siamese WSOD network using contrastive loss and fusion. By analyzing the relationship between language context and depth, we calculate depth priors to identify the bounding box proposals that may contain an object of interest. These depth priors are then utilized to update the list of pseudo ground-truth boxes, or adjust the confidence of per-box predictions. We evaluate our proposed method on three datasets (COCO, PASCAL VOC, and Conceptual Captions) by implementing it on top of two state-of-the-art WSOD methods, and we demonstrate a substantial enhancement in performance.
\end{abstract}

\input{intro}
\input{related}
\input{method}

\input{experiment}

\input{conclusion}

{\small
\bibliographystyle{ieee_fullname}
\bibliography{egbib}
}

\end{document}


\begin{appendices}

\renewcommand{\thesection}{S.\arabic{section}}
\renewcommand{\thetable}{S.\arabic{table}}
\renewcommand{\thefigure}{S.\arabic{figure}}

\title{Boosting Weakly Supervised Object Detection using Fusion and Priors from Hallucinated Depth\\(Supplementary File)}

\maketitle

\section{Alternative approach for depth priors}
\label{sec:alternate_approach}

\begin{table}[t]
\centering
\resizebox{\columnwidth}{!}{
\begin{tabular}{ lcccccc}
\toprule
 & \multicolumn{3}{c}{Avg. Precision, IoU}  & \multicolumn{3}{c}{Avg. Precision, Area} \\
 \cmidrule{2-4} \cmidrule{5-7}
Methods on COCO & 0.5:0.95 & 0.5 & 0.75 & S & M & L  \\
\midrule 
\textsc{MIST \cite{ren2020instance}} w/ EM &8.5&17.9&7.3 & 3.0 & 9.4 & 14.9 \\
\textsc{+ Siamese-Only} & 8.8 & 18.7 & 7.3 & 2.9 & 9.6 & 15.4  \\
\textsc{+ Depth-Oicr-Alt} & \underline{8.9} & \underline{19.0} & \underline{7.3} & \underline{3.0} & \underline{9.6} & \underline{15.6} \\
\textsc{+ Depth-Oicr} & \underline{9.0} & \underline{19.4} & \underline{7.3} & \underline{3.1} & \underline{9.6} & \underline{15.9} \\
\textsc{+ Depth-Attention-Alt} & \underline{9.0} & \underline{18.8} & \underline{7.7} & \underline{3.0} & 9.5 & \underline{16.0}  \\
\textsc{+ Depth-Attention} & \underline{9.1} & \underline{19.0} & \underline{7.9} & \underline{3.0} & 9.5 & \underline{16.2}  \\

\bottomrule
\hline \\
\end{tabular}}
\caption{This table introduces the comparison of the proposed and alternate approaches (\textsc{Depth-Oicr-Alt} and \textsc{Depth-Attention-Alt}) for modeling depth priors. All proposed methods that outperform the \textsc{Siamese-Only} are \underline{underlined}.}
\label{Table:alternate}
\end{table}

In Sec. 3.3, we described our method to obtain depth priors leveraging generated
bounding box predictions and associated captions to extract knowledge about the relative depths of objects. As an alternative, we use only bounding box predictions (without captions) to obtain depth priors.

Similar to Sec. 3.3, let $C$ be the set of object categories, and $B$ be the set of predicted bounding boxes. Let $d_{c,b} \in [0, 1]$ denote the depth value for object $c \in C$ and box $b \in B$. Further, $d_{c}$ represents a set of depth values for each $c$. The depth range $r_{c} = [mean - std, mean + std]$ is obtained by utilizing the mean and standard deviation (std) of this set of depth values in $d_{c}$. 

Table \ref{Table:alternate} demonstrates that the \textsc{Depth-Oicr-Alt} and \textsc{Depth-Attention-Alt} techniques, employed from the alternate approach, yield superior results compared to \textsc{Siamese-Only}. Nevertheless, the \textsc{Depth-Oicr} and \textsc{Depth-Attention} methods, derived from the proposed approach, outperform both \textsc{Depth-Oicr-Alt} and \textsc{Depth-Attention-Alt}. Note that the depth range $r_{c}$ remains consistent across all images in this alternative method. Conversely, in the proposed approach, the depth range $dr_{c}$ is computed individually for each image by taking into account the corresponding caption, as visualized in Figure 3 of the paper. As a result, the proposed approach demonstrates enhanced capacity in modeling depth priors through the utilization of captions. 

\section{Depth modality in inference}

As detailed in Section 3.2, fusion scores ($f^{det}$ and $f^{cls}$) are calculated by adding together RGB scores ($v^{det}$ and $v^{cls}$) and depth scores ($d^{det}$ and $d^{cls}$) in Eq. 5. Then, fusion scores are used to compute multiple instance learning loss $\mathcal{L}_{mil}$ in Eq. 9 to derive region-level scores during training. However, our proposed method uses only RGB detection $v^{det}$ and classification $v^{cls}$ scores during inference.

\begin{table}[t]
\centering
\resizebox{\columnwidth}{!}{
\begin{tabular}{ lcccccc}
\toprule
 & \multicolumn{3}{c}{Avg. Precision, IoU}  & \multicolumn{3}{c}{Avg. Precision, Area} \\
 \cmidrule{2-4} \cmidrule{5-7}

Methods on COCO & 0.5:0.95 & 0.5 & 0.75 & S & M & L  \\
\midrule 

\textsc{MIST \cite{ren2020instance}} & 11.8 & 24.3 & 10.7 & 3.6 & 13.2 & 18.9\\

\textsc{+ Wsod-Amplifier} (Ours) & \bf{13.8}&\bf{27.8}&\bf{12.5}&\bf{4.6}&\bf{14.8}&\bf{22.6}\\
\textsc{+ Wsod-Amplifier-Depth} & 4.4 & 8.7 & 3.8 & 0.6 & 3.5 & 8.8\\
\textsc{+ Wsod-Amplifier-Fusion} & 13.1 & 27.5 & 11.9 & 4.3 & 14.3 & 22.2\\

\bottomrule
\hline \\
\end{tabular}}
\caption{This table compares the different variations of our \textsc{Wsod-Amplifier} method during inference. The best performer per column is in \bf{bold}.}
\label{Table:inference}
\end{table}

The performance of \textsc{Wsod-Amplifier-Depth}, which relies solely on depth scores \emph{during inference}, is poorer due to the relatively lower informativeness of depth images compared to RGB images for detection in Table \ref{Table:inference}. On the other hand, \textsc{Wsod-Amplifier-Fusion}, which employs fusion scores, exhibits lower performance than our proposed approach, \textsc{Wsod-Amplifier}, which solely employs RGB scores during inference. The rationale behind the enhancement of results through fusion during training, while not during inference, stems from the nature of the MIL loss $\mathcal{L}_{mil}$. This loss function is formulated to guide the model in the classification of objects in an image, indirectly yielding region-level detection scores. While training, inaccuracies in depth for certain regions can be compensated by other regions to classify objects in MIL loss. However, during inference, each region operates independently, and errors on a per-region basis more directly impact detection results. Further, \cite{meyer2020improving} demonstrates that integrating the depth modality with RGB aids in enhancing the model's performance in classification tasks. The improved classification capacity of our model consequently leads to enhanced benefits from the MIL loss, contributing to improved representation learning. This improvement is highlighted by the enhanced performance of RGB scores during inference.

\input{table_per_class}

\section{Class-wise comparison on COCO}
Figure ~\ref{fig:coco_depth_ranges} demonstrates the effectiveness of depth priors in capturing the depth characteristics of different object classes in the COCO dataset. On average across classes, a substantial portion of the training data (80.75\%), resides within the constant ranges $r_{c}$ in the figure. Note the depth ranges $dr_{c}$ in our proposed approach for depth priors are computed separately for each image according to the corresponding caption. This image-specific calculation achieves more comprehensive coverage of depth variations within the training data compared to constant ranges $r_{c}$. Diverse object classes exhibit distinct depth ranges. The "fire hydrant" object displays the smallest average depth mean at 19\%, whereas the "kite" object displays the highest (54\%).


\section{Scalability and generalization}
Our demonstration reveals that a substantial amount of data is \emph{not} necessary for estimating depth. 
The depth priors calculated from the least frequent classes contribute even more to performance improvement than those from the most frequent classes. 
Referring to the data presented in Table \ref{Table:class-wise}, the average increase in $mAP_{50:95}$ for the least frequent 20\% classes is 3.4, while the increase for the most frequent 20\% classes is 0.8. 

To understand the impact of diverse depth ranges on performance, Table \ref{Table:class-wise} includes per-class standard deviations (STD). Notably, the depth priors computed from classes with smaller STDs make a more significant contribution to performance enhancement compared to those with larger STDs. On average, the $mAP_{50:95}$ increase for the 20\% of classes with the smallest STD is 4.8, whereas the increase for the 20\% of classes with larger STDs is 1.2. The observation suggests that a narrower depth distribution of an object corresponds to more informative depth information. Certainly, objects like "bear," "train," and "giraffe" exhibit smaller STDs, thereby leading to a more significant improvement in detection performance. Additionally, respective percentages of objects in Figure ~\ref{fig:coco_depth_ranges}, which depict the proportion of training data within the specified depth range, are notably higher at 89\%, 90\%, and 88\%, respectively.

Our analysis demonstrates that depth priors calculated using COCO exhibit similarities to those computed using PASCAL, as evident from Figure \ref{fig:voc_prior_two_plot}. On an average basis across various classes, approximately 84.4\% of data aligns within the constant ranges derived from PASCAL, whereas 82.3\% aligns within the ranges from COCO. The minimal differences in percentage and the visual resemblance of depth ranges in Figure \ref{fig:voc_prior_two_plot} underscore the generalizability of our depth priors. Furthermore, we mention in Sec. 4.3 that applying the depth priors calculated from COCO to Conceptual Captions (CC) results in a more significant performance improvement on the noisy CC compared to COCO, even though priors are computed from COCO. 

\input{table_domain_shift}

\begin{figure*}[t]
    \centering
    \includegraphics[width=\linewidth]{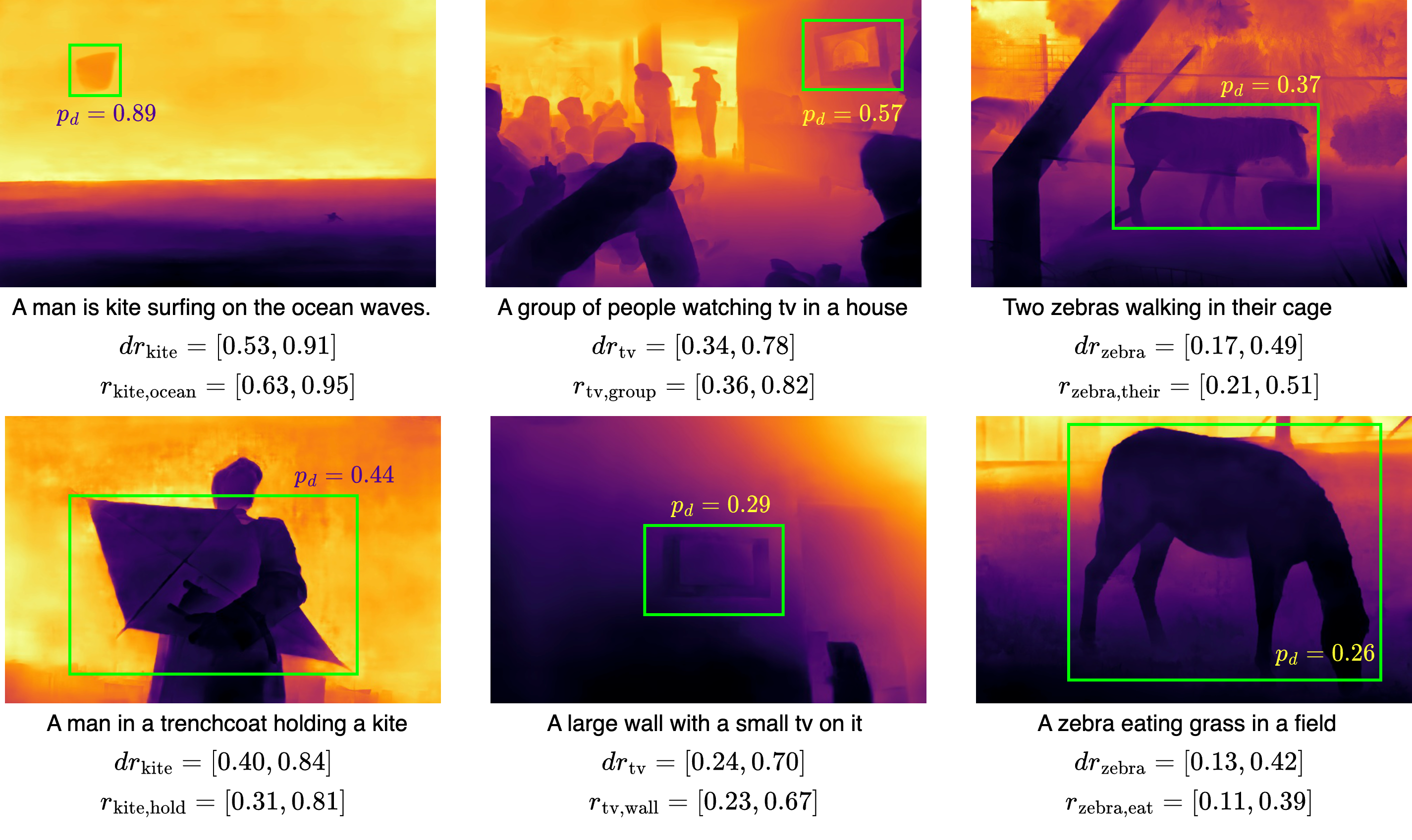}
    \caption{This figure provides an intuitive understanding of how the utilization of captions enhances the estimation of the depth range $dr_c$ of class c for a specific image. Here, $r_{c,w}$ signifies the depth range of class c when it is associated with the word w.}
    \label{fig:word_stat}
\end{figure*}

\section{Generalization to appearance changes}

By relying on depth information, our method builds some robustness to overfitting to appearance, which may not be the same across datasets. To test this hypothesis, we conduct experiments with domain shift datasets \cite{inoue2018cross}. Table ~\ref{Table:3} shows that our \textsc{Wsod-Amplifier} boosts the performance of MIST baseline in $mAP_{50}$ by $4-10\%$, even though no training is performed on these datasets.

\section{Illustration to depth ranges with caption}

The depth range $dr_c$ for class $c$ is individually computed for each image by leveraging the corresponding caption. In this context, $r_{c,w}$ denotes the depth range of class $c$ when linked to the word $w$. In Eq. 10, the calculation of $dr_c$ involves deriving the average of depth ranges $r_{c,w}$ across each word present within the caption. Figure \ref{fig:word_stat} illustrates three object examples: "kite", "TV", and "zebra". The top row of images depicts instances where objects are situated farther away from the camera, while the bottom row showcases examples where objects are positioned closer to the camera. As an illustration, consider the "kite" object in the upper image, where the caption features the word "ocean." Due to the influence of $r_{kite, ocean}$, the depth range $dr_{kite}$ becomes larger.  It's reasonable to expect that a kite would be positioned at a greater distance from the camera if it's flying over an ocean. Worth noting is that while $dr_{kite}$ is enlarged, it remains smaller than $r_{kite, ocean}$ due to the averaging effect with less descriptive words like ``is."  In a similar scenario, the depth range $dr_{kite}$ in the lower image becomes smaller due to the impact of $r_{kite, holding}$. When a person holds a kite, the likelihood is that the kite is positioned closer to the camera.

\section{Failure cases}
In Figure \ref{fig:failure}, the initial three cases depict objects with smaller depth values lying outside the depth range, while the last two cases feature objects with larger depth values. In the second scenario, the car is positioned closer to the camera, causing it to exceed the depth range limits. In contrast, in the fourth scenario, the broccoli is situated in the background with a larger depth value.

\begin{figure*}[t]
    \includegraphics[width=\textwidth]{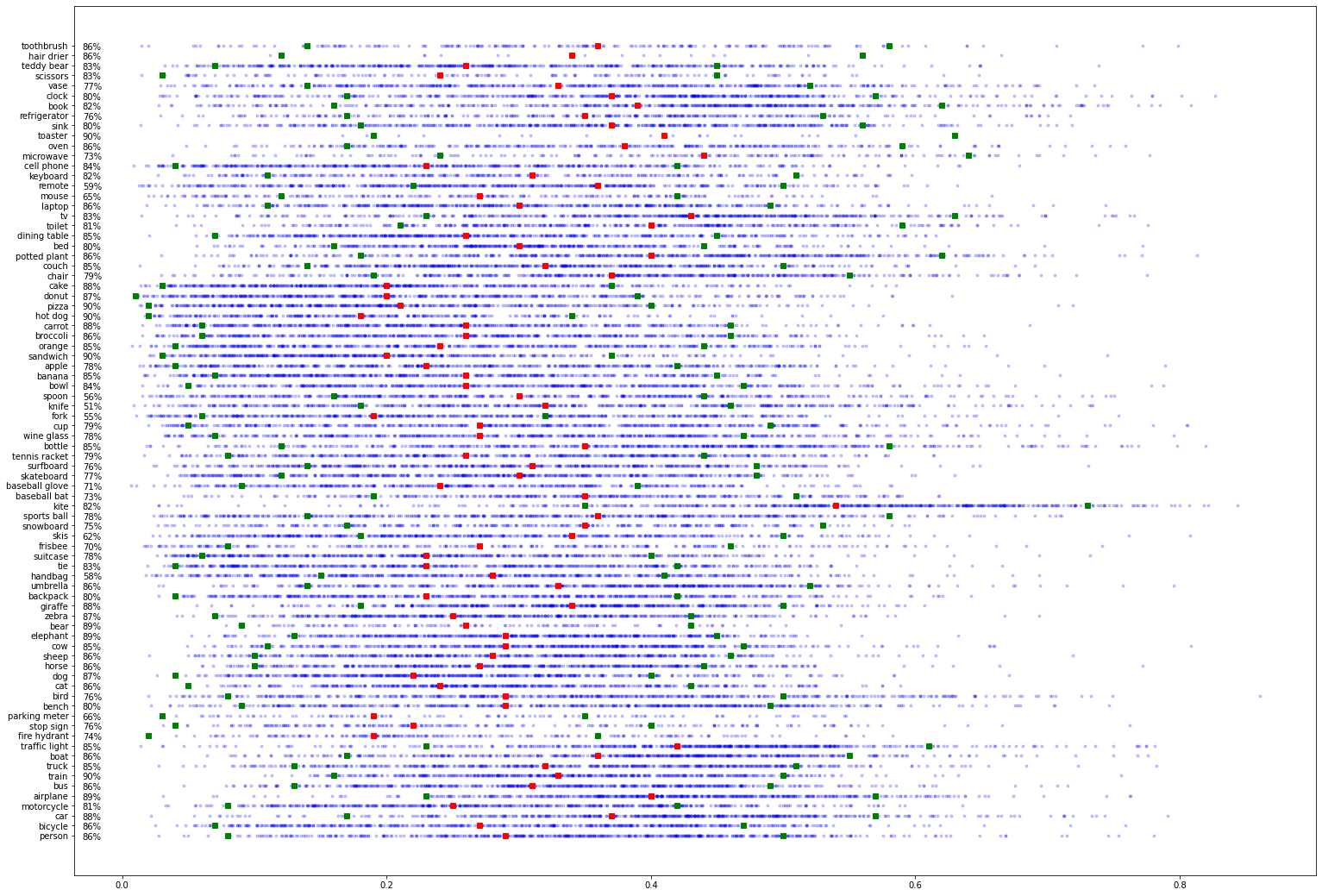}
    \caption{This figure illustrates how depth priors effectively capture depth characteristics across various object classes within the COCO dataset. In this visualization, points colored in blue indicate the depths of samples in the training set associated with the designated class.  Points colored in red denote the mean of depth range, and green points mark the boundaries of the range. The percentages located on the right side of class names provide insight into the proportion of training data falling within the defined depth range.}
 
    \label{fig:coco_depth_ranges}
\end{figure*}

\begin{figure*}[t]
    \includegraphics[width=\textwidth]{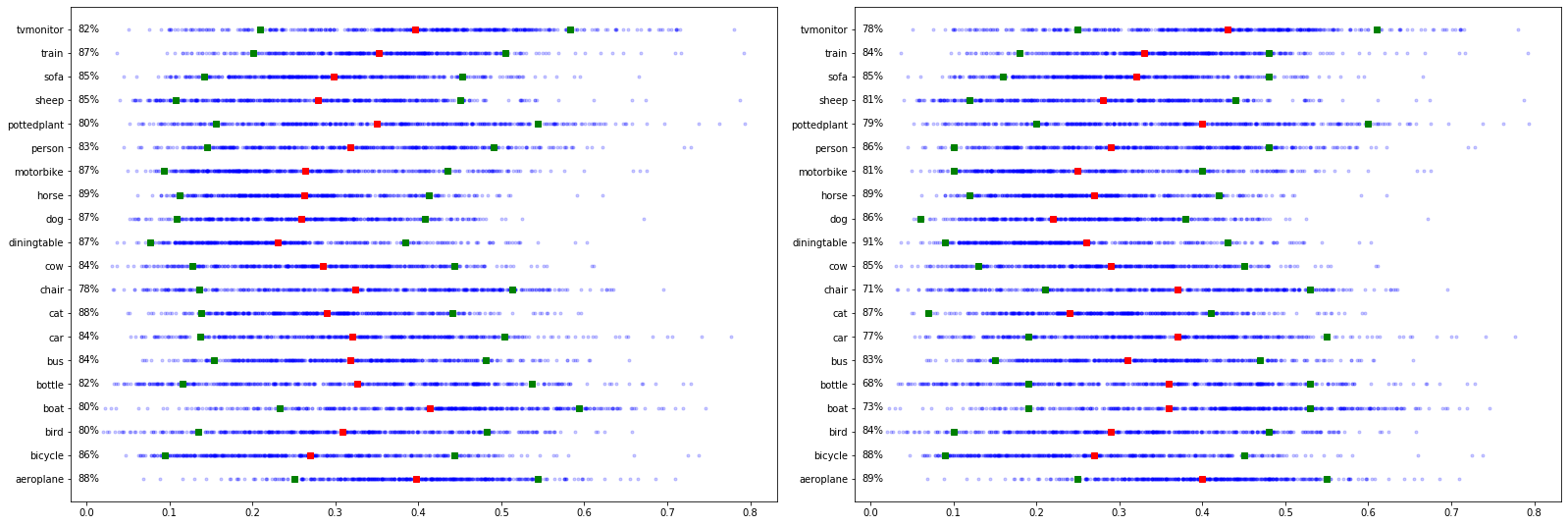}
    \caption{This figure demonstrates the effectiveness of depth priors, computed using the COCO dataset on the left and computed using the PASCAL dataset on the right, in capturing the depth characteristics across different object classes in the PASCAL dataset. }
 
    \label{fig:voc_prior_two_plot}
\end{figure*}

\begin{figure*}[t]
    \centering
    \includegraphics[width=\linewidth]{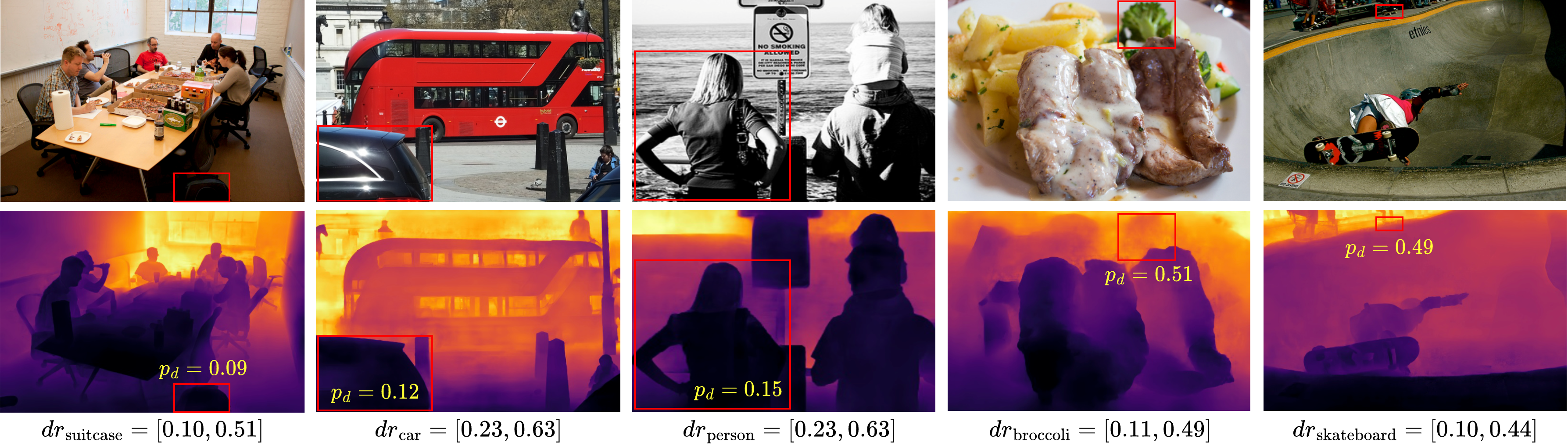}
    \caption{This figure illustrates the failure cases where the depth value $p_{d}$ of objects lies outside the depth range $dr_{c}$.}
    \label{fig:failure}
\end{figure*}

\clearpage
{\small
\bibliographystyle{ieee_fullname}
\bibliography{egbib}
}

\end{appendices}

%% file: intro.tex
 \section{Introduction}

\begin{figure}[t]
    \centering
    \includegraphics[width=\linewidth]{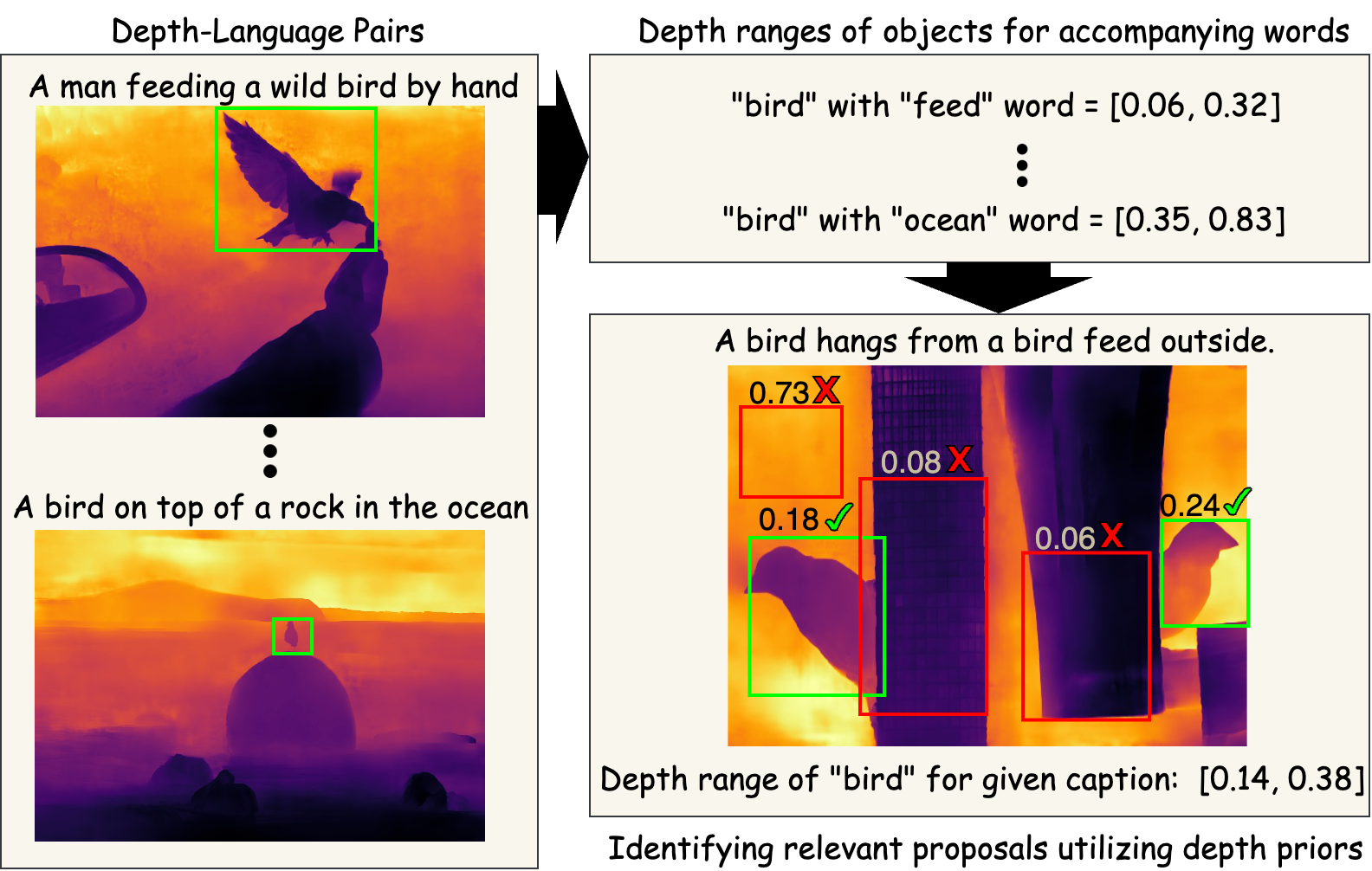}
    \caption{{Object from the same category may be at different depth depending on the context/setting. We use captions to capture context-conditioned depth ranges for each object class and co-occurring word: a bird may be closer when co-occurring with the word ``feed'' than the word ``ocean''.
    We use these ranges to spot relevant proposals that may contain target objects, and prune irrelevant ones, in weakly supervised object detection training. }}
    \label{fig:concept}
\end{figure}

Weakly-supervised object detection (WSOD) is a challenging task since it is unclear which instances have the label that was provided at the image level. Traditional methods only use appearance information in RGB images. However, appearance information is insufficient to localize objects in complex, cluttered environments. On the other hand, humans are capable of finding useful information in 
complex environments because they rely on object function, not just appearance.
For example, they might reason about which objects are within reach, which can be captured with depth from stereo vision \cite{cammack2016depth}. 
The depth modality provides additional cues about the spatial relationships and geometrical structure of objects in a scene and is invariant to appearance variations (e.g. in texture), making it complementary to the RGB modality. However, weakly-supervised object detection methods do not use depth information.

We equip WSOD methods with the ability to reason about functional information (depth). Importantly, our method does so without requiring additional annotations or suffering significant computational costs. We propose an amplifier method that can enhance the performance of different weakly supervised object detection methods based on multiple-instance learning. Since traditional WSOD datasets do not contain ground-truth depth information, the proposed method utilizes hallucinated (predicted) depth information obtained through a monocular depth estimation technique. During training, the method incorporates depth information to improve representation learning and to prune or down-weight predictions at the box level, which leads to improved object detection performance during inference.

First, depth can directly be used as a feature to aid representation learning, or to produce predictions which can be fused with those computed from RGB. While simple, this technique has not been used for WSOD, and we show that it is very effective: it boosts the performance of appearance-only methods by up to 2.6 mAP@0.5 (11\% relative gain).

Further, depth can provide strong priors about which of the bounding box proposals in the noisy WSOD setting contain an object of interest. We examine the rough depth at which objects of particular categories occur, by computing the depth range of an object using the predictions of a WSOD network. To make this range more precise, we examine the relationship between language context in captions and depth, by keeping track of depth range statistics conditioned on co-mentioned objects.
{The use of captions allows us to cope with the variable depth at which an object may occur depending on the context, as shown in Fig.~\ref{fig:concept}.}
We then use this range to prune the pseudo ground-truth bounding boxes used to iteratively update weakly-supervised detection methods, or to down-weight predictions at the box level.
This approach boosts WSOD performance further for a total up to 14\% mAP@0.5 gain.

Our method is simple and can boost multiple WSOD methods that rely on iterative improvement. We test it using two state-of-art WSOD baselines, MIST \cite{ren2020instance} and SoS-WSOD \cite{sui2022salvage}, on COCO and PASCAL VOC.
Inspired by recent work that trains object detection methods with language supervision \cite{zareian2021open,unal2022learning,gao}, we further test our method in a setting where labels at the image level are not ground-truth but estimated. In this setting, our method boosts the basic WSOD performance even more, by 18\% when labels for training are extracted from COCO, and 63\% when they are extracted from Conceptual Captions.

To summarize, our contributions are: 
    (1) We examine for the first time the use of depth in weakly-supervised object detection.
    (2) In addition to depth fusion, we propose a technique specific to WSOD, which estimates depth priors with the help of language, and uses them to refine pseudo boxes and box predictions. 
    (3) We show large performance gains in a large variety of settings, with the biggest boost from depth refinement when supervision is least expensive.





%% file: related.tex
\section{Related Work}
\textbf{Weakly-supervised object detection (WSOD)} is the task of learning to detect the location and type of objects given only image-level labels during training. The multi-instance learning (MIL) framework is commonly utilized in WSOD methods such as
WSDDN \cite{bilen2016weakly}.
OICR \cite{tang2017multiple} improved upon this by proposing pseudo-ground-truth mining and an online instance refinement, which was further refined by proposal clustering \cite{tang2018pcl}. C-MIL \cite{wan2019cmil} and MIST \cite{ren2020instance} introduced modifications to the MIL loss and pseudo-ground-truth mining, respectively. 
SoS-WSOD \cite{sui2022salvage} proposed a method that produces pseudo boxes for FSOD 
and splits noisy data for semi-supervised object detection.
Additionally, there have been efforts to bypass the need for image-level labels by utilizing noisy labels extracted from caption or subtitle data \cite{ ye2019cap2det,Chen_2017_CVPR,zareian2021open,unal2022learning,gao}.
Additionally, \cite{gungor2023complementary} leverages audio to improve WSOD performance and reduce noise from text-based label extraction.
In contrast to these works, our method leverages depth information as an additional modality, leading to improved performance in WSOD and a reduction of the noise in labels extracted from text.

\textbf{RGB-D detection.} 
The integration of RGB and depth information to derive complementary features has been previously studied for fully-supervised indoor analysis \cite{ying2022uctnet,zhou2022scale,jiao2019geometry,li2019mapnet, zhang2021depth} and object detection \cite{lee2022spsn,hussain2022pyramidal,feng2022encoder,fu2020jl, li2021joint, hoffman2016learning, huang2022good}. 
The strategies for merging the two modalities can be classified into three groups, depending on the point in the processing pipeline where the fusion occurs: early fusion \cite{fan2020rethinking,qu2017rgbd}, middle fusion \cite{fu2020jl,fu2021siamese, zhu2019pdnet, chen2018progressively}, and late fusion \cite{han2017cnns,piao2019depth}. Early fusion techniques involve combining the RGB and depth images into a single four-channel matrix at the earliest stage of the process. 
Middle fusion provides a balance between early and late fusion by utilizing CNNs for both feature extraction and subsequent merging. 
In late fusion, individual saliency prediction maps are produced from the RGB and depth channels to be combined through post-processing operations.  
In contrast to the majority of aforementioned methods, which use separate networks to extract features from RGB and depth images, several studies \cite{fu2020jl, fu2021siamese, song2021exploiting, meyer2020improving} employ Siamese networks to learn hierarchical features from both RGB and depth inputs by utilizing shared parameters.
However, \emph{we are the first to leverage depth data in weakly-supervised object detection.} Our approach is not specific to a particular method, as it can be applied to different MIL-based WSOD methods to improve their performance without incurring any extra annotation expenses and with minimal computational overhead during training. Although the depth modality is not used during the inference stage, incorporating it during training enhances the performance of the inference.

\textbf{Monocular depth estimation} involves predicting the depth map of a scene from a single RGB image \cite{miangoleh2021boosting,ranftl2020towards,ranftl2021vision,yin2021learning}.
We utilize the method in \cite{miangoleh2021boosting} to estimate depth on the training set due to its strong performance. 
This estimated (``hallucinated'') depth information is utilized to improve the performance of weakly supervised object detection.

%% file: method.tex
\section{Approach}

\begin{figure*}[t]
    \centering
    \includegraphics[width=1\linewidth]{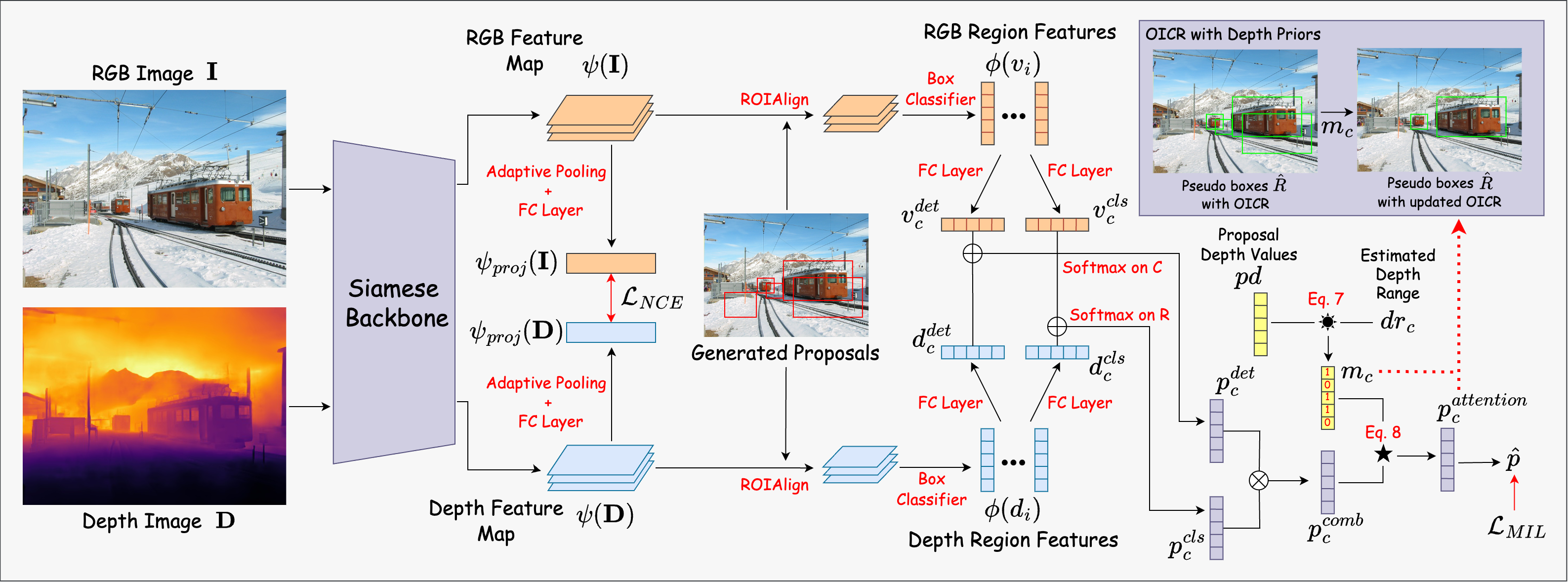}
    \caption{This figure illustrates the design of our proposed amplifier technique that takes advantage of depth information to enhance the performance of other weakly-supervised object detection methods. During inference, we only use the RGB branch (shown in orange).}
 
    \label{fig:main_fig}
\end{figure*}

We propose an amplifier approach that incorporates a depth modality to improve the effectiveness of WSOD methods.
Our method can be used with different MIL-based WSOD methods to boost their performance by incurring little extra cost during training.
It does not use the depth modality during inference to avoid any slow-downs and reliance on additional data (depth estimation or captions). The proposed approach comprises three main steps (Sec.~\ref{sec:siamese_module}, \ref{sec:fusion}, \ref{sec:depth}, respectively). First, a Siamese network with a shared backbone is employed to improve representation learning through contrastive learning between RGB and depth features (referred to as \textsc{Siamese-Only} in the experiments). Second, we combine detection and classification scores obtained from both the RGB and depth modalities, which can be categorized as late fusion (\textsc{Fusion}). Third, we use captions and bounding box predictions of traditional WSOD to calculate depth priors.
These depth priors are then used to improve the OICR-style \cite{tang2017multiple} module in two WSOD methods (named \textsc{Depth-OICR}) and create attention with combined score probabilities (\textsc{Depth-Attention}).
Note that \textsc{Siamese-Only} is always applied, while \textsc{Fusion}, \textsc{Depth-OICR} and \textsc{Depth-Attention} build on top of it, and can be used alone or combined.

\subsection{The Siamese WSOD Network}
\label{sec:siamese_module}

\textbf{WSOD.}
Following Bilen et al. \cite{bilen2016weakly}, let $\mathbf{I} \in \mathbb{R}^{h\times w \times 3}$ denote an RGB image, and $y_c \in \{0, 1\}$ (where $c \in \{1, . . . , C\}$ and $C$ is the total number of object categories) be its corresponding ground-truth class labels.
Let $v_i$, $i \in \{1, . . . , R\}$ (where $R$ is the number of proposals), 
denote the visual proposals in image $\mathbf{I}$. 
RoI pooling is applied and a fixed-length feature vector $\phi(v_i)$ extracted for each visual region. The proposal features $\phi(v_i)$ are fed into two parallel fully-connected layers to compute the visual detection score $v^{det}_{i,c} \in \mathbb{R}^1$ and classification score $v^{cls}_{i,c} \in \mathbb{R}^1$: 

\begin{equation}
   v^{det}_{i,c} = w^{det\intercal}_{c} \phi(v_i) + b^{det}_{c}, \quad v^{cls}_{i,c} = w^{cls\intercal}_{c} \phi(v_i) + b^{cls}_{c}
   \label{eq:1}
\end{equation}
where $w$ and $b$ are weights and bias, 
respectively.

\textbf{Estimating the depth images.} To extract depth information from RGB images, we employ the monocular depth estimation technique by Mahdi et al. \cite{miangoleh2021boosting}. This enables us to use existing RGB-only object detection datasets without the need for additional annotations. Although the extracted depth images are initially grayscale, we use a color map to convert them to RGB images with three channels.

\textbf{Siamese design.}
Our approach utilizes a Siamese network with contrastive learning to incorporate depth information in the weakly-supervised object detection network during training. 
This design allows us to use a backbone pre-trained with RGB images to extract features from both RGB and depth images, without adding extra complexity to the model's parameters. We enhance the representation learning of the backbone by defining contrastive loss between RGB and depth features similar to \cite{meyer2020improving}.
Utilizing a Siamese network provides the advantage of using only RGB images during inference similar to other WSOD methods. This ensures that our contribution does not introduce any additional overhead on the inference time.

With the help of a pre-trained backbone model, the feature map of RGB image $\psi(\mathbf{I})$ is extracted. Let $\mathbf{D} \in \mathbb{R}^{h\times w \times 3}$ denote a depth image associated with the RGB image $\mathbf{I}$ and let $\psi(\mathbf{D})$ be the feature map of the depth image $\mathbf{D}$ extracted by the Siamese backbone. The RGB feature map $\psi(\mathbf{I})$ and depth feature map $\psi(\mathbf{D})$ are fed into adaptive pooling and fully connected layers to obtain $d$-dimensional projected feature vectors $\psi_{proj}(\mathbf{I})$ and $\psi_{proj}(\mathbf{D})$.
The only extra parameters we add to the traditional MIL-based WSOD network come from the fully connected layer for projection with 8 percent overhead (13M parameters for the projection layer, vs 154M total). 
If no late fusion is performed in the experiments, we train as described in Sec.~\ref{sec:fusion}, but excluding the $d_{i, c}$ variables in Eq.~\ref{eq:fusion}. 

\textbf{Contrastive learning.}
We L2-normalize the RGB and depth feature vectors $\psi_{proj}(\mathbf{I})$ and $\psi_{proj}(\mathbf{D})$ vectors, and compute their cosine similarity:
\begin{equation}
    S(\mathbf{I},\mathbf{D}) = 	\langle \psi_{proj}(\mathbf{I}), \psi_{proj}(\mathbf{D}) \rangle / \rho
    \label{eq:b}
\end{equation} where $\rho$, is a learnable temperature parameter.
We use noise contrastive estimation (NCE) \cite{gutmann2010noise} to define the contrastive learning by considering RGB image and depth image pairs $(\mathbf{I}, \mathbf{D}) \in \mathcal{B}$ where $\mathcal{B}$ is an RGB-depth pair batch. The first component of the NCE loss contrasts an RGB image with negative depth images to measure how closely the RGB image matches with its paired depth among others in the batch:

\begin{equation}
\resizebox{\columnwidth}{!}{$
\mathcal{L}_{{D}\rightarrow{I}} = -\dfrac{1}{|\mathcal{B}|} \sum_{(\mathbf{I},\mathbf{D}) \in \mathcal{B}} log \dfrac{exp(S(\mathbf{I},\mathbf{D}))}{exp(S(\mathbf{I},\mathbf{D})) + \sum_{(\mathbf{I}',\mathbf{D}') \in \mathcal{B}} exp(S(\mathbf{I},\mathbf{D}'))}$}
\end{equation} 

The second component of the NCE loss, $\mathcal{L}_{{I} \rightarrow {D}}$, is analogously defined to contrast a depth image with negative RGB image samples, and
the two components are averaged:
\begin{equation}
    \mathcal{L}_{NCE} = (\mathcal{L}_{{D} \rightarrow {I}} + \mathcal{L}_{{I} \rightarrow {D}}) / 2
\end{equation}

\subsection{Late Fusion of the Modalities}
\label{sec:fusion}

The detection and classification scores computed from RGB and depth modalities are imbued with disparate and complementary details that jointly enrich our understanding of the target objects. Therefore, we combine these scores to amplify the performance of object detection.

As the depth images are derived from the RGB images, the spatial arrangement of the objects is equivalent in both modalities. Hence, we utilize the same visual region proposals for both RGB and depth modalities. Following the application of the RoI pooling layer and the Siamese box feature extractor to the depth feature map $\psi(\mathbf{D})$, we obtain the feature vector $\phi(d_i)$ for each depth region. Thereafter, we employ the approach presented in Eq.~\ref{eq:1} to derive the depth detection score $d^{det}_{i,c} \in \mathbb{R}^1$ and the depth classification score $d^{cls}_{i,c} \in \mathbb{R}^1$. Subsequently, we fuse (sum) the scores from the RGB and depth modalities:
\begin{equation}
  f^{det}_{i,c} = v^{det}_{i,c} + d^{det}_{i,c}, \quad f^{cls}_{i,c} = v^{cls}_{i,c} + d^{cls}_{i,c}
\label{eq:fusion}
\end{equation}
where $f^{det}_{i,c}$ and $f^{cls}_{i,c}$ are fusion detection and classification scores, respectively.

Following the WSDDN \cite{bilen2016weakly} architecture, these classification and detection scores are converted to probabilities such that $p^{cls}_{i,c}$ is the probability that class $c$ is in present proposal $f_i$, and $p^{det}_{i,c}$ is the probability that $f_i$ is important for predicting image-level label $y_c$.

\begin{equation}
   p^{det}_{i,c} = \dfrac{exp(f^{det}_{i,c})} {\sum_{k=1}^{R} exp(f^{det}_{k,c})}, \quad p^{cls}_{i,c} = \dfrac{exp(f^{cls}_{i,c})} {\sum_{k=1}^{C} exp(f^{cls}_{i,k})}
   \label{eq:7}
\end{equation}

We element-wise multiply the classification and detection scores to obtain the combined score $p^{comb}_{i,c}$:
\begin{equation}
  p^{comb}_{i,c} = p^{det}_{i,c} p^{cls}_{i,c}
  \label{eq:8}
\end{equation}

Finally, image-level predictions $\hat{p}_c$ are computed as follows, where greater values of $\hat{p}_c\in [0, 1]$ mean a higher likelihood that $c$ is present in the image.
\begin{equation}
  \hat{p}_c = \sigma\left(\sum_{i=1}^{R}p^{comb}_{i,c} \right)
  \label{eq:confidence}
\end{equation}
Assuming the label $y_c$ = 1 if and only if class $c$ is present, the classification loss used for training the model is defined as follows. Since no region-level labels are provided, we must derive region-level scores indirectly, by optimizing this loss.
\begin{equation}
    \mathcal{L}_{mil} = -\sum_{c=1}^{C}\left[ y_c\log\hat{p}_c + (1-y_c)\log(1-\hat{p}_c) \right]
    \label{eq:visual_loss}
\end{equation}

\subsection{Depth Priors}
\label{sec:depth}

We utilize the baseline WSOD methods, which we aim to improve, to generate bounding box object predictions in the training set. Further, we leverage both the generated bounding box predictions and associated captions to extract knowledge about the relative depths of objects. We note that our proposed methodology adheres to the WSOD setting, deriving benefits from the predicted bounding boxes, as opposed to ground truth bounding box annotations to calculate depth priors. We subsequently exploit these depth priors to guide the identification of the relevant visual regions that may contain the target objects.
Further, we show that even though we estimate the depth priors from COCO, they generalize to Conceptual Captions (Table \ref{Table:2}).

We use the notation $pd_i \in [0, 1]$, $i \in {1, . . . , R}$, where $R$ is the number of pre-computed region proposals for depth image $\mathbf{D}$, to represent the average depth value in the $i$-th region proposal. Each region proposal contains pixels with values ranging from 0 to 1, which correspond to the smallest and largest depth values, respectively. 

\begin{figure}[t]
\includegraphics[width= \columnwidth]{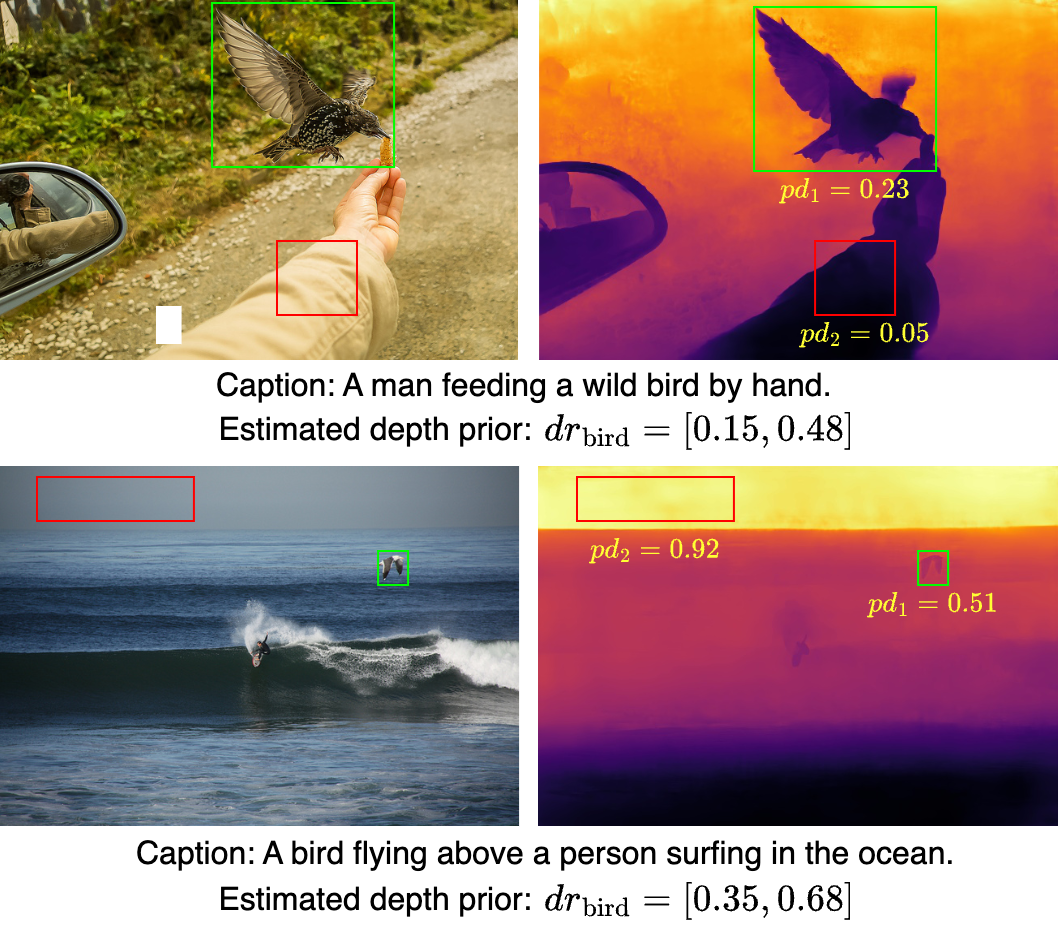} 
\caption{The figure displays a row of images that are accompanied by their respective depth and caption data, as well as proposal depth value of different regions and estimated depth prior range.}
\label{fig:prior}
\end{figure}

We employ bounding box predictions $B$ to approximate the depth value of objects using the caption that describes the image in which the objects are present. 
{We also use co-occurring captions to capture the context in which an object occurs, and condition depth priors on this context which varies across images.}
Let $C$ be the set of object categories, $W$ be the set of distinct words in the vocabulary that includes every word in the captions, and $B$ be the set of predicted bounding boxes. Let $d_{c,w,b} \in \{[0, 1], \varnothing\}$ denote the depth value for object $c \in C$, word $w \in W$ and box $b \in B$, which is calculated by averaging the depth values in the pixels of $b$ similar to the calculation of $pd_i$.  As an example, $d_{\text{bird},\text{ocean}, b}$ represents the depth value of the ``bird" object box $b$ of a depth image that has a caption that includes the ``ocean" word. In the absence of ``ocean" in the caption or when annotation $b$ does not correspond to the ``bird" object, the depth value $d_{\text{bird},\text{ocean}, b}$ is set to null $\varnothing$. Further, $d_{c,w}$ represents a set of depth values calculated {by averaging $d_{c,w,b}$ over predicted boxes $b \in B$,} excluding $\varnothing$ values. The depth range $r_{c,w} = [mean - std, mean + std]$ for class $c$ and word $w$ is obtained by utilizing the mean and standard deviation (std) of this set of depth values in $d_{c,w}$. 
Once these depth ranges $r_{c, w}$ are computed, they can be applied to estimate an allowable depth range for a class $c$ in a new image, without any boxes on that image.


For any new depth image $\mathbf{D}$, the range of estimated depth priors for an object $c$ is $dr_c$:
\begin{equation}
dr_{c} = \dfrac{\sum_{s \in S}{r_{c,s}}}{|S|}
\label{eq:prior}
\end{equation}
where $S$ denotes the set of words in the caption corresponding to $\mathbf{D}$. {Thus the depth at which we expect to find objects in a particular image varies depending on the context provided by words in the corresponding caption.}
We \textbf{only require captions at training time.}

We utilize the estimated depth prior range $dr_c$ to identify potentially important regions in $pd$ for each class. We define a depth mask indicator variable $m_{i,c} \in {0,1}$ for each region $i$ $\in$ $R$ and class $c$ $\in$ $C$, which indicates the likelihood of a particular region in an image containing an object of a certain class. The computation of this variable is as follows:
\begin{equation}
    m_{i,c}= 
\begin{cases}
    1,& \text{if } pd_i \in dr_c\\
    0,              & \text{otherwise}
\end{cases}
\label{eq:mask}
\end{equation}
If the proposal depth value $pd_i$ falls within the estimated depth prior range $dr_c$ for class $c$, it is considered as a relevant region for that class, and the corresponding mask variable $m_{i,c}$ is set to 1; otherwise, it is set to 0. Subsequently, we utilize the mask variable $m_{i,c}$ in combination with our end-to-end network to improve its performance.

As an example, Fig.~\ref{fig:prior} presents two images featuring a ``bird" object, with different depths. The estimated depth prior ranges $dr_\text{bird}$
are calculated using Eq.~\ref{eq:prior} for each image based on the words in the caption. The caption of the first image includes ``feeding" and ``hand" which suggest the ``bird" is likely to have a smaller depth, while the caption of the second image includes ``flying" and ``ocean" that suggest the ``bird" is likely to have a bigger depth. The regions on the images having a proposal depth value of $pd_1$ are in the estimated depth prior range $dr_\text{bird}$; we observe that they truly include the ``bird" object. 
The range allows us to rule out regions with values $pd_2$, which do not contain ``bird''.

{ 
\textbf{Alternative method for estimating depth priors.}
As an alternative, we use only bounding box predictions (without captions) to obtain depth priors. 
Let $d_{c,b} \in [0, 1]$ denote the depth value for object $c \in C$ and box $b \in B$. Further, 
let $d_{c}$ represent a set of depth values for each $c$. The depth range $r_{c} = [mean - std, mean + std]$ is obtained by utilizing the mean and standard deviation (std) of this set of depth values in $d_{c}$. Then we set $dr_c = r_c$ as we do not use caption information (compared to Eq.~\ref{eq:prior}); $dr_c$ is used in Eq.~\ref{eq:mask}.
}

\subsubsection{Depth Priors: Updated OICR}
\label{sec:depth_oicr}
\newcommand\sForAll[2]{ \ForAll{#1}#2\EndFor} 
\newcommand\sIf[2]{ \If{#1}#2\EndIf}
\begin{algorithm}
	\caption{OICR Mining with Depth Priors} 
        Input: Proposals $R$, Depth Mask Indicator Variable $m$\\
        Output: Pseudo boxes $\hat{R}$
	\begin{algorithmic}[1]
        \State $\hat{R} = \varnothing$ 
        \sForAll {$c=1:C$}{
            \sForAll {$i=1:|R|$}{
                \State $\hat{R_c} = \hat{R_c} \cup R_i $ \textbf{if} $m_{i,c} = 1$
            }
        }
        \State \textbf{return} $\hat{R}$
	\end{algorithmic} 
\label{algorithm:algo}
\end{algorithm}

 Online Instance Classifier Refinement (OICR) \cite{tang2017multiple} is a weakly supervised object detection algorithm that iteratively refines object proposals.  Recent studies \cite{tang2018pcl,ren2020instance,sui2022salvage} have highlighted the importance of more effective proposal mining strategies for achieving better recall and precision of objects in WSOD detectors. We propose an algorithm that incorporates the depth priors during the proposal mining provided in Alg.~\ref{algorithm:algo}. As our proposed method aims to enhance MIL-based WSOD methods, we utilize our algorithm in conjunction with recent OICR-style/self-training/mining strategies, subject to the depth prior condition specified in the fourth line of Alg.~\ref{algorithm:algo}. After using the depth prior condition, OICR-style mining selects fewer but more relevant proposals so our contribution increases mining precision.\footnote{In early experiments, we verified our method's gains persist if the baseline 
 drops the lowest-scoring pseudo boxes without using depth.} 

 \subsubsection{Depth Priors: Attention}
 \label{sec:depth_attention}
The depth mask variable $m_{i,c}$ indicates the potentially important proposal regions for each class. We use this variable to employ an attention mechanism with combined score probabilities $p^{comb}_{i,c}$ provided in Eq. ~\ref{eq:8} as follows:
\begin{equation}
    p^{comb}_{i,c}= 
    p^{comb}_{i,c} * 0.5, \text{if } m_{i,c} = 0
\end{equation}
This mechanism reduces the probability of a region for class $c$ by half if the region is determined as less likely to be important by $m_{i,c}$. These scores are then used in Eq.~\ref{eq:confidence}.

%% file: experiment.tex
\section{Experiments}

We test our method on top of two weakly-supervised detection techniques, and verify the contributions of  the constituents of our approach:

\begin{itemize}[nolistsep,noitemsep]
\item Siamese WSOD Network (\textsc{Siamese-Only}, Sec.~\ref{sec:siamese_module});
\item Late Fusion of the Modalities (\textsc{Fusion}, Sec.~\ref{sec:fusion}) which combines classification/detection from RGB and depth, and builds on top of the Siamese WSOD Network (Sec.~\ref{sec:siamese_module});
\item Depth Priors are utilized to enhance the OICR-style module (\textsc{Depth-Oicr}, Sec.~\ref{sec:depth_oicr}) and construct attention (\textsc{Depth-Attention}, Sec.~\ref{sec:depth_attention}) with visual-only score probabilities, both building upon the Siamese WSOD Network (Sec.~\ref{sec:siamese_module}); 
\item Finally, we use all components of our method (\textsc{Wsod-Amplifier, Sec.~\ref{sec:siamese_module}, \ref{sec:fusion}, \ref{sec:depth_oicr}, \ref{sec:depth_attention}}).
\item {\textsc{Depth-Oicr-Alt} and \textsc{Depth-Attention-Alt} estimate depth priors without captions.}
\item {\textsc{Wsod-Amplifier-Inf} fuses RGB and depth at inference time, unlike our proposed method.}
\end{itemize}

\subsection{Experimental Setup}


\textbf{PASCAL Visual Object Classes 2007 (VOC-07)} \cite{everingham2009pascal} contains 20 classes. For training, we use 2501 images from the train set and 2510 images from the validation set. We evaluate using 4952 images from the test set.

\textbf{Common Objects in Context (COCO)} \cite{lin2014microsoft} consists of 80 classes. 
We utilize approximately 118k images from the train set and use the labels provided at the image level. Additionally, to test how well our method works when labels are obtained from noisy language supervision (in captions), we train our models using labels obtained through an exact match (EM) method following \cite{unal2022learning}, also referred to as substring matching in \cite{fang2022data}. Due to the unavailability of any labels for around 15k images extracted from captions, we excluded them from the training set and use 103k images. We evaluate using 5k images from the validation set.

\textbf{Conceptual Captions (CC)} \cite{sharma2018conceptual} is a large-scale image captioning dataset containing over 3 million images annotated with only captions. We use around 30k images and their corresponding captions and the labels are extracted for the 80 COCO dataset classes using an exact match method from the captions. During the evaluation, we used 5k images from the COCO validation set.

\textbf{Domain shift datasets.} 
{In the supplementary}, we also evaluate our method in a domain shift setting \cite{inoue2018cross}, using three datasets. Clipart1k has the same 20 classes as VOC with 1,000 images, while Watercolor2k and Comic2k share 6 classes with VOC and have 2,000 images each. 

\textbf{Evaluation protocols.} We utilize mean Average Precision (mAP) considering various IoU thresholds as the common evaluation metric for COCO and VOC datasets. Additionally, we report mAP for objects of different sizes during COCO evaluation and we report the results of Correct Localization (CorLoc) for VOC evaluation.

\textbf{Implementation details.}
We employ the official PyTorch implementations of SoS-WSOD \cite{sui2022salvage} and MIST \cite{ren2020instance} methods to apply our amplifier technique. SoS-WSOD uses four images per GPU as two augmented images and their flipped versions with a total of 4 GPUs, whereas MIST uses only one image per GPU with a total of 8 GPUs. However, we use one image per GPU for SoS-WSOD due to VRAM limitation in our GPUs, as we also utilized depth images for each corresponding RGB image. Therefore, the baseline results of SoS-WSOD reported in Table ~\ref{Table:1} are slightly lower than those reported in the original paper. Moreover, we solely use the first stage of SoS-WSOD since it includes the MIL-based WSOD module which is convenient to implement our method on top of. The other settings are kept the same as the official implementations with the VGG16 backbone. The inference is done on the training set by using baseline MIST and SoS-WSOD methods to obtain box predictions having confidence scores higher than 0.5. These box predictions are then used to calculate depth priors. Furthermore, we utilize the same depth range $r_{c,w}$ from the COCO annotations for the \textsc{Wsod-Amplifier} method on the Conceptual Captions dataset. 

\subsection{Comparing our amplifier to state of the art}

\input{table_gt_coco_voc}

We evaluate our proposed methods, \textsc{Fusion} and \textsc{Wsod-Amplifier}, using two state-of-the-art WSOD approaches, SoS-WSOD \cite{sui2022salvage} and MIST \cite{ren2020instance}, and the COCO and VOC-07 datasets. The performance of our proposed methods are compared with the baseline methods in Table ~\ref{Table:1}. 
When our \textsc{Wsod-Amplifier} method is applied to MIST, it improves the baseline performance by $17\%$ in $mAP_{50:95}$ (relative gain, 13.8/11.8-1) and $14\%$ in $mAP_{50}$. Similarly, when our \textsc{Wsod-Amplifier} method is applied to SoS-WSOD, it improves the baseline performance by $2\%$ in $mAP_{50:95}$, $1.5\%$ in $mAP_{50}$, and $6\%$ in $mAP_{75}$. As the VOC-07 dataset does not have 
captions, we are only able to apply the \textsc{Siamese-Only} and \textsc{Fusion} methods on SoS-WSOD but not the \textsc{Depth-Oicr} and \textsc{Depth-Attention}. On this dataset, our improvements outperformed the baseline SoS-WSOD by $5\%$ in $mAP_{50:95}$, $2\%$ in $mAP_{50}$, and $9\%$ in $mAP_{75}$. 
{\textsc{Wsod-Amplifier-Inf} performs worse than \textsc{Wsod-Amplifier}. We argue depth is useful as a soft guide to balance region information during MIL training, but less so when directly used in the strict detection setting.}

\input{table_em_coco_cc}

\subsection{Ablation studies and visualization}
\textbf{Experiments with labels from captions.} Several attempts \cite{ye2019cap2det, unal2022learning,fang2022data} have been made to eliminate the requirement for image-level labels by leveraging noisy label information obtained from captions or subtitles. Although it is cost-effective to use text information for label extraction, it results in a decrease in the performance of weakly supervised object detection. \cite{unal2022learning} propose a text classifier approach to extract labels more effectively than the simple exact match (EM) and reduce the noise between text and ground truth (GT) labels. In contrast to previous studies, our research employs the depth modality to reduce the noise in labels extracted from captions. Our approach improves the model's detection capability and employs captions during the calculation of depth priors. We conducted experiments with MIST \cite{ren2020instance} using both GT and EM labels and observed that, as expected, training with GT labels leads to significantly better performance than training with EM labels in Table ~\ref{Table:2} due to the noise in labels extracted from captions. However, our proposed \textsc{Wsod-Amplifier} method applied on MIST with EM labels surpasses the baseline and MIST with GT labels. These findings demonstrate that our method \emph{effectively reduces noise} and enables the model trained with EM labels to achieve better performance than those trained with GT labels. It is worth noting that the text classifier approach proposed by \cite{unal2022learning} also performs better than EM-labeled training data, but falls short of the performance achieved by GT-labeled data.

\textbf{Results on noisy datasets.}
We also extract labels from captions on the Conceptual Captions dataset, which lacks labels at the image level. We observe that our \textsc{Wsod-Amplifier} boosts results by an impressive 63\% relative gain using $mAP_{50}$. 
Conceptual Captions is a noisier dataset than COCO, since captions were not collected through crowdsourcing, but were crawled as alt-text for web search results. 
Thus, it is noteworthy that \emph{the benefit of our approach becomes more pronounced as the cost of supervision decreases, and the noise in the supervision increases.}

\begin{figure*}[t]
    \centering
    \includegraphics[width=\linewidth]{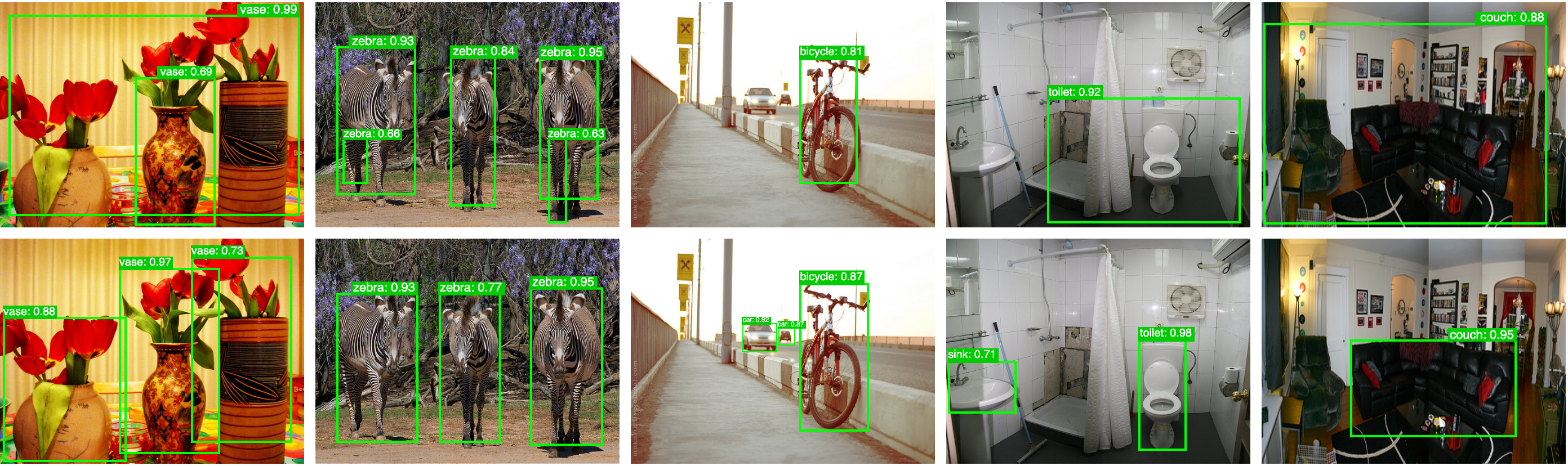}
    \caption{Qualitative comparison of MIST \cite{ren2020instance} (top) and our proposed \textsc{Wsod-Amplifier} method (bottom),  
    on COCO val. The ground-truth objects are vase, zebra, bicycle and car, sink and toilet, couch in this order. Confidence scores and names of objects shown. }
    \label{fig:qualitative}
\end{figure*}

\textbf{Analysing the components of our approach.} To understand the impact of each component of our approach on the overall performance, we conducted experiments with MIST \cite{ren2020instance} using EM labels as a baseline and applied each component of our method on top of the baseline in Table ~\ref{Table:2}. Our \textsc{Siamese-Only} method, which incorporates the depth modality in the Siamese network using contrastive learning, improves feature extraction and results in a $4\%$ increase in $mAP_{50:95}$ and $mAP_{50}$. Our \textsc{Depth-Oicr} method, which utilizes depth priors in the OICR module to improve the mining strategy, increases $mAP_{50:95}$ and $mAP_{50}$ over \textsc{Siamese-Only} by $6-8\%$ on COCO and $42-47\%$ on Conceptual Captions (CC). Our \textsc{Depth-Attention} method, which incorporates depth priors to use potentially important regions in an attention mechanism with combined score probabilities, increases $AP_{50:95}$ and $mAP_{75}$ by $6-7\%$. Our \textsc{Fusion} method, which combines RGB and depth image scores, improves by $14-16\%$ on COCO and $7-21\%$ on CC. 
Comparing \textsc{Fusion} and \textsc{Depth-Oicr}, the bigger gain using $mAP_{50}$ is achieved by \textsc{Fusion} on COCO, and \textsc{Depth-Oicr} on CC. \emph{Thus, the benefit of our WSOD-specific method component increases as the noise in the dataset increases, which is appealing due to its real-world applicability.} 
Finally, our \textsc{Wsod-Amplifier} method, which includes all components of our approach, achieves the highest performance increase over MIST w/ EM baseline, with \emph{improvements in all $mAP$ metrics by $16-20\%$ on COCO and $53-65\%$ on CC.} 

{
\textbf{Alternative depth priors.} 
At the bottom of Table \ref{Table:2} (COCO), we see that \textsc{Depth-Oicr-Alt} and \textsc{Depth-Attention-Alt}, derived from the alternative approach, yield superior results compared to \textsc{Siamese-Only}. However, the \textsc{Depth-Oicr} and \textsc{Depth-Attention} methods, derived from our full (caption-based) approach, outperform both \textsc{Depth-Oicr-Alt} and \textsc{Depth-Attention-Alt}. Note that the depth range $r_{c}$ remains consistent across all images in this alternative method. Conversely, in the proposed approach, the depth range $dr_{c}$ is computed individually for each image by taking into account the corresponding caption, as visualized in Fig.~\ref{fig:prior}. As a result, the proposed approach demonstrates enhanced capacity in modeling depth priors through the utilization of captions. }

\textbf{Generalization of depth priors.} {Even though on CC we use the depth priors calculated from COCO}, our proposed method exhibits a more substantial enhancement in CC performance compared to COCO, achieving an improvement of $63\%$ $mAP_{50}$ over the MIST baseline (50\% improvement from \textsc{Depth-Oicr} alone). 
{Thus, our \textsc{Depth-Oicr} demonstrates generalization, as it has a higher impact than \textsc{Fusion} on CC (without recomputing priors), in contrast to COCO.}
Given the recent interest in learning from vision-language data, our approach has the potential to be highly impactful.
{Further, we compared the priors estimated from different datasets, and found them to be similar. In particular, 82.3\% of PASCAL objects fit within the $[mean-stdev, mean+stdev]$ range computed from COCO, and 84.4\% when the range is computed on PASCAL itself; the cross-domain gap in the range is small.}



\textbf{Qualitative analysis.} We visualize the object detection performance of our proposed \textsc{Wsod-Amplifier} compared to MIST \cite{ren2020instance} in Fig.~~\ref{fig:qualitative}. 
The confidence scores are calculated using visual detection $v^{det}$ and classification scores $v^{cls}$. We show boxes with scores higher than 0.5. In the first image, the baseline struggles to accurately identify multiple instances of the same ``vase" objects, instead grouping them together in a single box. 
Our method overcomes this challenge, precisely detecting each individual ``vase". In the second image, the baseline faces the problem of part domination due to some discriminative parts of a ``zebra". Our method overcomes this issue by utilizing depth modality during training, which emphasizes the geometric variations of objects, while comparatively ignoring the complex background.
In other images, unlike our method, the baseline misses objects entirely, or produces large and imprecise bounding boxes.
Moreover, the boxes detected by our method tend to have higher prediction scores.

%% file: table_gt_coco_voc.tex
\begin{table}[t]
\centering
\resizebox{\columnwidth}{!}{
\begin{tabu}{ lcccccc}
\toprule
 & \multicolumn{3}{c}{Avg. Precision, IoU}  & \multicolumn{3}{c}{Avg. Precision, Area} \\
 \cmidrule{2-4} \cmidrule{5-7}
Methods on COCO & 0.5:0.95 & 0.5  & 0.75 & S & M  & L  \\

\midrule 

\textsc{MIST \cite{ren2020instance}} & 11.8 & 24.3 & 10.7 & 3.6 & 13.2 & 18.9\\
\textsc{+ Fusion} & 13.5&26.9&{12.4}&4.0&14.7&21.6\\
\textsc{+ Wsod-Amplifier} & \bf{13.8}&\bf{27.8}&\bf{12.5}&\bf{4.6}&\bf{14.8}&\bf{22.6}\\
\textsc{+ Wsod-Amplifier-Inf} & 13.1 & 27.5 & 11.9 & 4.3 & 14.3 & 22.2\\

\midrule 
\textsc{SoS-WSOD \cite{sui2022salvage}} & 10.2 & 21.5 & 8.6 & \bf{2.5} & 10.6 & 17.7\\
\textsc{+ Fusion} & 10.3 & 21.6 & 8.9 & 2.3 & 10.8 & 18.4\\
\textsc{+ Wsod-Amplifier} & \bf{10.5} & \bf{21.8} & \bf{9.1} &  \bf{2.5} & \bf{11.1} & \bf{18.7} \\

\midrule
\midrule
 & \multicolumn{3}{c}{Avg. Precision, IoU}  & \multicolumn{3}{c}{CorLoc} \\
 \cmidrule{2-4} \cmidrule{5-7}
Methods on VOC-07 &  0.5:0.95 & 0.5  & 0.75  & 0.5:0.95 & 0.5  & 0.75 \\
\midrule 
\textsc{SoS-WSOD \cite{sui2022salvage}} & 24.8&52.2&20.4&38.7&71.7&36.9\\
\textsc{+ Fusion} & \bf{26.0}&\bf{53.1}&\bf{22.3}&\bf{39.6}&\bf{72.1}&\bf{38.5}\\

\bottomrule
\hline \\
\end{tabu}}
\caption{This table compares the performance enhancement of our methods, to their baseline results SoS-WSOD \cite{sui2022salvage} and MIST \cite{ren2020instance}, on COCO and VOC-07. The best performer per column is in \textbf{bold}.} 
\label{Table:1}
\end{table}

%% file: table_em_coco_cc.tex
\begin{table}[t]
\centering
\resizebox{\columnwidth}{!}{
\begin{tabu}{ lcccccc}
\toprule
 & \multicolumn{3}{c}{Avg. Precision, IoU}  & \multicolumn{3}{c}{Avg. Precision, Area} \\
 \cmidrule{2-4} \cmidrule{5-7}

Methods on COCO & 0.5:0.95 & 0.5 & 0.75 & S & M & L  \\

\midrule 
\textsc{MIST \cite{ren2020instance}} w/ GT &9.7&21.1&8.0 & 3.0 & 10.4 & 15.1 \\
\midrule[0.1pt]
\textsc{MIST \cite{ren2020instance}} w/ EM &8.5&17.9&7.3 & 3.0 & 9.4 & 14.9 \\
\textsc{+ Siamese-Only} & 8.8 & 18.7 & 7.3 & 2.9 & 9.6 & 15.4  \\
\textsc{+ Depth-Oicr} & \underline{9.0} & \underline{19.4} & \underline{7.3} & \underline{3.1} & \underline{9.6} & \underline{15.9} \\
\textsc{+ Depth-Attention} & \underline{9.1} & \underline{19.0} & \underline{7.9} & \underline{3.0} & 9.5 & \underline{16.2}  \\
\textsc{+ Fusion} & \underline{9.9} & \underline{20.4} & \underline{8.5} & \underline{3.0} & \underline{10.1} & \underline{17.1} \\
\textsc{+ Wsod-Amplifier} & \bf{\underline{10.2}} & \bf{\underline{21.0}} & \bf{\underline{8.5}} & \bf{\underline{3.3}} & \bf{\underline{10.3}} & \bf{\underline{17.5}}  \\
\midrule[0.1pt]
  \textsc{+ Depth-Oicr-Alt} & \underline{8.9} & \underline{19.0} & \underline{7.3} & \underline{3.0} & \underline{9.6} & \underline{15.6} \\
  \textsc{+ Depth-Attention-Alt} & \underline{9.0} & \underline{18.8} & \underline{7.7} & \underline{3.0} & 9.5 & \underline{16.0}  \\

\midrule
\midrule
 & \multicolumn{3}{c}{Avg. Precision, IoU}  & \multicolumn{3}{c}{Avg. Precision, Area} \\
 \cmidrule{2-4} \cmidrule{5-7}

Methods on CC & 0.5:0.95 & 0.5 & 0.75 & S & M & L  \\
\midrule 
\textsc{MIST \cite{ren2020instance}} w/ EM  &1.7&3.8&1.4 & 0.3 & 1.7 & 3.4 \\

\textsc{+ Fusion} & 2.0 & {4.1} & {1.7} & {0.3} & {1.9} & {4.0} \\

\textsc{+ Depth-Oicr} & {2.4} & {5.6} & {{2.0}} & {0.3} & {2.2} & {5.1} \\
\textsc{+ Wsod-Amplifier} & \bf{{2.6}} & \bf{{6.3}} & \bf{2.1} & \bf{{0.4}} & \bf{{2.8}} & \bf{{5.7}}  \\

\bottomrule
\hline \\
\end{tabu}}

\caption{This table introduces the effect of each component of our method implemented on MIST \cite{ren2020instance} with exact match (EM) labels on COCO (top) and Conceptual Captions (CC) (bottom). The best performer per column is in \textbf{bold}. On top, all proposed methods that outperform the \textsc{Siamese-Only} are \underline{underlined}.} 
\label{Table:2}
\end{table}

%% file: conclusion.tex
\textbf{Conclusion.}
We show depth boosts weakly-supervised object detection methods, tested on SoS-WSOD and MIST, without extra annotation or costly computation. 
Our Siamese WSOD network efficiently incorporates RGB and depth with contrastive learning and fusion. Using the relation of language and depth, depth priors estimate the bounding box proposals that may contain an object of interest.

\textbf{Acknowledgement:} This work was supported by a National Science Foundation Award No. 2046853, and a University of Pittsburgh Momentum Funds award. 

%% file: table_per_class.tex
\begin{table}[t]
\centering
\resizebox{0.94\columnwidth}{!}{
\begin{tabular}{ lcccc}
\toprule

Objects & \# of Instances & STD & MIST \cite{ren2020instance} w/ EM  & \textsc{+ Wsod-Amplifier} \\
\midrule 
Bear&902&15&19.9&36.6 (\textcolor{blue}{+16.7})\\
Toilet&2860&17&4.7&20.3 (\textcolor{blue}{+15.6})\\
Train&3157&15&14.3&26.3 (\textcolor{blue}{+12.0})	\\
TV&4031&18&11.3&19.2 (\textcolor{blue}{+7.9})	\\
Giraffe&3593&14&8.0&13.6 (\textcolor{blue}{+5.6})	\\
Zebra&3653&16&22.4&27.9 (\textcolor{blue}{+5.5})	\\
Airplane&3823&15&28.7&33.9 (\textcolor{blue}{+5.2})	\\
Elephant&3876&14&26.4&31.1 (\textcolor{blue}{+4.7})	\\
Refrigerator&1872&16&8.8&13.5 (\textcolor{blue}{+4.7})	\\
Cat&3298&17&7.8&11.8 (\textcolor{blue}{+4.0})	\\
Pizza&3993&17&32.2&35.9 (\textcolor{blue}{+3.7})	\\
Horse&4645&15&21.5&25.1 (\textcolor{blue}{+3.6})	\\
Fire hydrant&1313&15&24.8&28.4 (\textcolor{blue}{+3.6})	\\
Sheep&6442&16&11.3&14.5 (\textcolor{blue}{+3.2})	\\
Cow&5588&16&19.7&22.8 (\textcolor{blue}{+3.1})	\\
Dog&3764&16&8.7&11.7 (\textcolor{blue}{+3})	\\
Truck&7043&17&9.7&12.6 (\textcolor{blue}{+2.9})	\\
Cake&4508&15&6.2&9.0 (\textcolor{blue}{+2.8})	\\
Parking meter&833&14&23.7&26.4 (\textcolor{blue}{+2.7})	\\
Sandwich&3069&15&8.0&10.6 (\textcolor{blue}{+2.6})	\\
Bird&7100&19&8.8&11.3 (\textcolor{blue}{+2.5})	\\
Kite&6333&17&11.2&13.6 (\textcolor{blue}{+2.4})	\\
Carrot&5463&18&1.7&4.0 (\textcolor{blue}{+2.3})	\\
Broccoli&4894&18&5.3&7.6 (\textcolor{blue}{+2.3})	\\
Clock&4310&18&13.0&15.3 (\textcolor{blue}{+2.3})	\\
Motorcycle&5971&15&15.4&17.5 (\textcolor{blue}{+2.1})	\\
Tennis racket&3397&16&1.7&3.7 (\textcolor{blue}{+2.0})	\\
Keyboard&1978&18&1.4&3.3 (\textcolor{blue}{+1.9})	\\
Donut&4854&17&12.3&14.0 (\textcolor{blue}{+1.7})	\\
Cell phone&4449&17&11.5&13.1 (\textcolor{blue}{+1.6})	\\
Apple&4244&17&5.0&6.5 (\textcolor{blue}{+1.5})	\\
Car&30530&18&6.1&7.6 (\textcolor{blue}{+1.5})	\\
Banana&6698&17&5.6&7.1 (\textcolor{blue}{+1.5})	\\
Chair&26825&16&1.5&2.8 (\textcolor{blue}{+1.3})	\\
Hot dog&1997&14&7.8&9.0 (\textcolor{blue}{+1.2})	\\
Umbrella&7729&17&6.9&8.1 (\textcolor{blue}{+1.2})	\\
Wine glass&5559&18&0.4&1.5 (\textcolor{blue}{+1.1})	\\
Sink&3929&17&0.1&1.2 (\textcolor{blue}{+1.1})	\\
Bench&6739&18&6.1&7.0 (\textcolor{blue}{+0.9})	\\
Microwave&1188&18&0.8&1.6 (\textcolor{blue}{+0.8})	\\
Bed&2903&12&27.6&28.4 (\textcolor{blue}{+0.8})	\\
Vase&4593&17&0.7&1.4 (\textcolor{blue}{+0.7})	\\
Bowl&10020&19&11.7&12.4 (\textcolor{blue}{+0.7})	\\
Bus&4317&16&35.5&36.2 (\textcolor{blue}{+0.7})	\\
Laptop&3406&17&30.2&30.8 (\textcolor{blue}{+0.6})	\\
Bottle&16782&21&4.6&5.2 (\textcolor{blue}{+0.6})	\\
Boat&7449&17&4.2&4.7 (\textcolor{blue}{+0.5})	\\
Bicycle&4912&18&9.8&10.3 (\textcolor{blue}{+0.5})	\\
Teddy bear&3397&17&17.6&17.9 (\textcolor{blue}{+0.3})	\\
Cup&14454&20&3.8&4.0 (\textcolor{blue}{+0.2})	\\
Snowboard&1956&16&0.0&0.2 (\textcolor{blue}{+0.2})	\\
Surfboard&4133&15&0.2&0.4 (\textcolor{blue}{+0.2})	\\
Stop sign&1372&16&48.0&48.2 (\textcolor{blue}{+0.2})	\\
Couch&4111&16&1.2&1.3 (\textcolor{blue}{+0.1})	\\
Book&16826&21&0.1&0.2 (\textcolor{blue}{+0.1})	\\
Scissors&1059&19&0.1&0.1 \\
Skis&4683&14&0.1&0.0 (\textcolor{red}{-0.1})	\\
Suitcase&4188&15&3.8&3.6 (\textcolor{red}{-0.2})	\\
Person&181524&19&1.4&1.0 (\textcolor{red}{-0.4})	\\
Traffic light&9115&17&5.9&4.7 (\textcolor{red}{-1.2})	\\
Frisbee&1861&17&6.8&4.8 (\textcolor{red}{-2.0})	\\
Orange&4525&18&13.1&10.6 (\textcolor{red}{-2.5})	\\
Oven&2302&19&9.7&5.7 (\textcolor{red}{-4.0})\\
\midrule
Total&597245&15&8.5&10.2 (\textcolor{blue}{+1.7})\\
\bottomrule
\hline \\
\end{tabular}}
\caption{The table provides class-wise $mAP_{50:95}$ results for both MIST \cite{ren2020instance} with EM and our \textsc{Wsod-Amplifier}. The objects are sorted according to the change in performance.  Additionally, the table provides information about the standard deviation (STD) and the number of instances within the training set. Note that 17 object categories from the COCO dataset have been excluded from this table due to both methods yielding a $mAP$ of zero.} 
\label{Table:class-wise}
\end{table}


%% file: table_domain_shift.tex
\begin{table}[t]
\centering
\resizebox{0.8\columnwidth}{!}{
\begin{tabular}{ lccc}
\toprule

Methods & Clipart & Watercolor & Comic  \\
\midrule 
\textsc{MIST \cite{ren2020instance} w/ GT}  &9.4&13.3&9.2  \\

\textsc{+ Wsod-Amplifier} & \bf{{10.2}} & \bf{{14.7}} & \bf{9.6}   \\
\bottomrule
\hline \\
\end{tabular}}
\caption{This table introduces the improvement of our \textsc{Wsod-Amplifier} over MIST on domain shift datasets. The results are in $mAP_{50}$ metric. The best performer per column is in \textbf{bold}.} 
\label{Table:3}
\end{table}

%% file: PaperForReview.bbl
\begin{thebibliography}{10}\itemsep=-1pt

\bibitem{bilen2016weakly}
Hakan Bilen and Andrea Vedaldi.
\newblock Weakly supervised deep detection networks.
\newblock In {\em Proceedings of the IEEE/CVF Conference on Computer Vision and
  Pattern Recognition (CVPR)}, pages 2846--2854, 2016.

\bibitem{cammack2016depth}
P Cammack and JM Harris.
\newblock Depth perception in disparity-defined objects: finding the balance
  between averaging and segregation.
\newblock {\em Philosophical Transactions of the Royal Society B: Biological
  Sciences}, 371(1697):20150258, 2016.

\bibitem{chen2018progressively}
Hao Chen and Youfu Li.
\newblock Progressively complementarity-aware fusion network for rgb-d salient
  object detection.
\newblock In {\em Proceedings of the IEEE/CVF Conference on Computer Vision and
  Pattern Recognition (CVPR)}, pages 3051--3060, 2018.

\bibitem{Chen_2017_CVPR}
Kai Chen, Hang Song, Chen Change~Loy, and Dahua Lin.
\newblock Discover and learn new objects from documentaries.
\newblock In {\em Proceedings of the IEEE/CVF Conference on Computer Vision and
  Pattern Recognition (CVPR)}, 2017.

\bibitem{everingham2009pascal}
Mark Everingham, Luc Van~Gool, Christopher~KI Williams, John Winn, and Andrew
  Zisserman.
\newblock The pascal visual object classes (voc) challenge.
\newblock {\em International Journal of Computer Vision}, 88:303--308, 2009.

\bibitem{fan2020rethinking}
Deng-Ping Fan, Zheng Lin, Zhao Zhang, Menglong Zhu, and Ming-Ming Cheng.
\newblock Rethinking rgb-d salient object detection: Models, data sets, and
  large-scale benchmarks.
\newblock {\em IEEE Transactions on Neural Networks and Learning Systems},
  32(5):2075--2089, 2020.

\bibitem{fang2022data}
Alex Fang, Gabriel Ilharco, Mitchell Wortsman, Yuhao Wan, Vaishaal Shankar,
  Achal Dave, and Ludwig Schmidt.
\newblock Data determines distributional robustness in contrastive language
  image pre-training (clip).
\newblock In {\em International Conference on Machine Learning (ICML)}, pages
  6216--6234. PMLR, 2022.

\bibitem{feng2022encoder}
Guang Feng, Jinyu Meng, Lihe Zhang, and Huchuan Lu.
\newblock Encoder deep interleaved network with multi-scale aggregation for
  rgb-d salient object detection.
\newblock {\em Pattern Recognition}, 128:108666, 2022.

\bibitem{fu2020jl}
Keren Fu, Deng-Ping Fan, Ge-Peng Ji, and Qijun Zhao.
\newblock Jl-dcf: Joint learning and densely-cooperative fusion framework for
  rgb-d salient object detection.
\newblock In {\em Proceedings of the IEEE/CVF Conference on Computer Vision and
  Pattern Recognition (CVPR)}, pages 3052--3062, 2020.

\bibitem{fu2021siamese}
Keren Fu, Deng-Ping Fan, Ge-Peng Ji, Qijun Zhao, Jianbing Shen, and Ce Zhu.
\newblock Siamese network for rgb-d salient object detection and beyond.
\newblock {\em IEEE Transactions on Pattern Analysis and Machine Intelligence
  (TPAMI)}, 44(9):5541--5559, 2021.

\bibitem{gao}
Mingfei Gao, Chen Xing, Juan~Carlos Niebles, Junnan Li, Ran Xu, Wenhao Liu, and
  Caiming Xiong.
\newblock Open vocabulary object detection with pseudo bounding-box labels.
\newblock In {\em European Conference on Computer Vision (ECCV)}, 2022.

\bibitem{gungor2023complementary}
Cagri Gungor and Adriana Kovashka.
\newblock Complementary cues from audio help combat noise in weakly-supervised
  object detection.
\newblock In {\em Proceedings of the IEEE/CVF Winter Conference on Applications
  of Computer Vision (WACV)}, pages 2185--2194, 2023.

\bibitem{gutmann2010noise}
Michael Gutmann and Aapo Hyv{\"a}rinen.
\newblock Noise-contrastive estimation: A new estimation principle for
  unnormalized statistical models.
\newblock In {\em Proceedings of the thirteenth international conference on
  artificial intelligence and statistics}, pages 297--304. JMLR Workshop and
  Conference Proceedings, 2010.

\bibitem{han2017cnns}
Junwei Han, Hao Chen, Nian Liu, Chenggang Yan, and Xuelong Li.
\newblock Cnns-based rgb-d saliency detection via cross-view transfer and
  multiview fusion.
\newblock {\em IEEE Transactions on Cybernetics}, 48(11):3171--3183, 2017.

\bibitem{hoffman2016learning}
Judy Hoffman, Saurabh Gupta, and Trevor Darrell.
\newblock Learning with side information through modality hallucination.
\newblock In {\em Proceedings of the IEEE/CVF Conference on Computer Vision and
  Pattern Recognition (CVPR)}, pages 826--834, 2016.

\bibitem{huang2022good}
Haiwen Huang, Andreas Geiger, and Dan Zhang.
\newblock Good: Exploring geometric cues for detecting objects in an open
  world.
\newblock In {\em International Conference on Learning Representations (ICLR)},
  2023.

\bibitem{hussain2022pyramidal}
Tanveer Hussain, Abbas Anwar, Saeed Anwar, Lars Petersson, and Sung~Wook Baik.
\newblock Pyramidal attention for saliency detection.
\newblock In {\em 2022 IEEE/CVF Conference on Computer Vision and Pattern
  Recognition Workshops (CVPRW)}, pages 2877--2887. IEEE, 2022.

\bibitem{inoue2018cross}
Naoto Inoue, Ryosuke Furuta, Toshihiko Yamasaki, and Kiyoharu Aizawa.
\newblock Cross-domain weakly-supervised object detection through progressive
  domain adaptation.
\newblock In {\em Proceedings of the IEEE/CVF Conference on Computer Vision and
  Pattern Recognition (CVPR)}, pages 5001--5009, 2018.

\bibitem{jiao2019geometry}
Jianbo Jiao, Yunchao Wei, Zequn Jie, Honghui Shi, Rynson~WH Lau, and Thomas~S
  Huang.
\newblock Geometry-aware distillation for indoor semantic segmentation.
\newblock In {\em Proceedings of the IEEE/CVF Conference on Computer Vision and
  Pattern Recognition (CVPR)}, pages 2869--2878, 2019.

\bibitem{lee2022spsn}
Minhyeok Lee, Chaewon Park, Suhwan Cho, and Sangyoun Lee.
\newblock Spsn: Superpixel prototype sampling network for rgb-d salient object
  detection.
\newblock In {\em Computer Vision--ECCV 2022: 17th European Conference, Tel
  Aviv, Israel, October 23--27, 2022, Proceedings, Part XXIX}, pages 630--647.
  Springer, 2022.

\bibitem{li2021joint}
Jingjing Li, Wei Ji, Qi Bi, Cheng Yan, Miao Zhang, Yongri Piao, Huchuan Lu,
  et~al.
\newblock Joint semantic mining for weakly supervised rgb-d salient object
  detection.
\newblock {\em Advances in Neural Information Processing Systems},
  34:11945--11959, 2021.

\bibitem{li2019mapnet}
Yabei Li, Zhang Zhang, Yanhua Cheng, Liang Wang, and Tieniu Tan.
\newblock Mapnet: Multi-modal attentive pooling network for rgb-d indoor scene
  classification.
\newblock {\em Pattern Recognition}, 90:436--449, 2019.

\bibitem{lin2014microsoft}
Tsung-Yi Lin, Michael Maire, Serge Belongie, James Hays, Pietro Perona, Deva
  Ramanan, Piotr Doll{\'a}r, and C~Lawrence Zitnick.
\newblock Microsoft coco: Common objects in context.
\newblock In {\em Computer Vision--ECCV 2014: 13th European Conference, Zurich,
  Switzerland, September 6-12, 2014, Proceedings, Part V 13}, pages 740--755.
  Springer, 2014.

\bibitem{meyer2020improving}
Johannes Meyer, Andreas Eitel, Thomas Brox, and Wolfram Burgard.
\newblock Improving unimodal object recognition with multimodal contrastive
  learning.
\newblock In {\em 2020 IEEE/RSJ International Conference on Intelligent Robots
  and Systems (IROS)}, pages 5656--5663. IEEE, 2020.

\bibitem{miangoleh2021boosting}
S~Mahdi~H Miangoleh, Sebastian Dille, Long Mai, Sylvain Paris, and Yagiz Aksoy.
\newblock Boosting monocular depth estimation models to high-resolution via
  content-adaptive multi-resolution merging.
\newblock In {\em Proceedings of the IEEE/CVF Conference on Computer Vision and
  Pattern Recognition (CVPR)}, pages 9685--9694, 2021.

\bibitem{piao2019depth}
Yongri Piao, Wei Ji, Jingjing Li, Miao Zhang, and Huchuan Lu.
\newblock Depth-induced multi-scale recurrent attention network for saliency
  detection.
\newblock In {\em Proceedings of the IEEE/CVF International Conference on
  Computer Vision (ICCV)}, pages 7254--7263, 2019.

\bibitem{qu2017rgbd}
Liangqiong Qu, Shengfeng He, Jiawei Zhang, Jiandong Tian, Yandong Tang, and
  Qingxiong Yang.
\newblock Rgbd salient object detection via deep fusion.
\newblock {\em IEEE Transactions on Image Processing}, 26(5):2274--2285, 2017.

\bibitem{ranftl2021vision}
Ren{\'e} Ranftl, Alexey Bochkovskiy, and Vladlen Koltun.
\newblock Vision transformers for dense prediction.
\newblock In {\em Proceedings of the IEEE/CVF International Conference on
  Computer Vision (ICCV)}, pages 12179--12188, 2021.

\bibitem{ranftl2020towards}
Ren{\'e} Ranftl, Katrin Lasinger, David Hafner, Konrad Schindler, and Vladlen
  Koltun.
\newblock Towards robust monocular depth estimation: Mixing datasets for
  zero-shot cross-dataset transfer.
\newblock {\em IEEE Transactions on Pattern Analysis and Machine Intelligence
  (TPAMI)}, 44(3):1623--1637, 2020.

\bibitem{ren2020instance}
Zhongzheng Ren, Zhiding Yu, Xiaodong Yang, Ming-Yu Liu, Yong~Jae Lee,
  Alexander~G Schwing, and Jan Kautz.
\newblock Instance-aware, context-focused, and memory-efficient weakly
  supervised object detection.
\newblock In {\em Proceedings of the IEEE/CVF Conference on Computer Vision and
  Pattern Recognition (CVPR)}, 2020.

\bibitem{sharma2018conceptual}
Piyush Sharma, Nan Ding, Sebastian Goodman, and Radu Soricut.
\newblock Conceptual captions: A cleaned, hypernymed, image alt-text dataset
  for automatic image captioning.
\newblock In {\em Proceedings of the 56th Annual Meeting of the Association for
  Computational Linguistics (Volume 1: Long Papers)}, pages 2556--2565, 2018.

\bibitem{song2021exploiting}
Hwanjun Song, Eunyoung Kim, Varun Jampan, Deqing Sun, Jae-Gil Lee, and
  Ming-Hsuan Yang.
\newblock Exploiting scene depth for object detection with multimodal
  transformers.
\newblock In {\em 32nd British Machine Vision Conference (BMVC)}, pages 1--14.
  British Machine Vision Association (BMVA), 2021.

\bibitem{sui2022salvage}
Lin Sui, Chen-Lin Zhang, and Jianxin Wu.
\newblock Salvage of supervision in weakly supervised object detection.
\newblock In {\em Proceedings of the IEEE/CVF Conference on Computer Vision and
  Pattern Recognition (CVPR)}, pages 14227--14236, 2022.

\bibitem{tang2018pcl}
Peng Tang, Xinggang Wang, Song Bai, Wei Shen, Xiang Bai, Wenyu Liu, and Alan
  Yuille.
\newblock Pcl: Proposal cluster learning for weakly supervised object
  detection.
\newblock {\em IEEE Transactions on Pattern Analysis and Machine Intelligence
  (TPAMI)}, 42(1):176--191, 2018.

\bibitem{tang2017multiple}
Peng Tang, Xinggang Wang, Xiang Bai, and Wenyu Liu.
\newblock Multiple instance detection network with online instance classifier
  refinement.
\newblock In {\em Proceedings of the IEEE/CVF Conference on Computer Vision and
  Pattern Recognition (CVPR)}, 2017.

\bibitem{unal2022learning}
Mesut~Erhan Unal, Keren Ye, Mingda Zhang, Christopher Thomas, Adriana Kovashka,
  Wei Li, Danfeng Qin, and Jesse Berent.
\newblock Learning to overcome noise in weak caption supervision for object
  detection.
\newblock {\em IEEE Transactions on Pattern Analysis and Machine Intelligence
  (TPAMI)}, 2022.

\bibitem{wan2019cmil}
Fang Wan, Chang Liu, Wei Ke, Xiangyang Ji, Jianbin Jiao, and Qixiang Ye.
\newblock C-mil: Continuation multiple instance learning for weakly supervised
  object detection.
\newblock In {\em Proceedings of the IEEE/CVF Conference on Computer Vision and
  Pattern Recognition (CVPR)}, 2019.

\bibitem{ye2019cap2det}
Keren Ye, Mingda Zhang, Adriana Kovashka, Wei Li, Danfeng Qin, and Jesse
  Berent.
\newblock Cap2det: Learning to amplify weak caption supervision for object
  detection.
\newblock In {\em Proceedings of the IEEE/CVF International Conference on
  Computer Vision (CVPR)}, pages 9686--9695, 2019.

\bibitem{yin2021learning}
Wei Yin, Jianming Zhang, Oliver Wang, Simon Niklaus, Long Mai, Simon Chen, and
  Chunhua Shen.
\newblock Learning to recover 3d scene shape from a single image.
\newblock In {\em Proceedings of the IEEE/CVF Conference on Computer Vision and
  Pattern Recognition (CVPR)}, pages 204--213, 2021.

\bibitem{ying2022uctnet}
Xiaowen Ying and Mooi~Choo Chuah.
\newblock Uctnet: Uncertainty-aware cross-modal transformer network for indoor
  rgb-d semantic segmentation.
\newblock In {\em Computer Vision--ECCV 2022: 17th European Conference, Tel
  Aviv, Israel, October 23--27, 2022, Proceedings, Part XXX}, pages 20--37.
  Springer, 2022.

\bibitem{zareian2021open}
Alireza Zareian, Kevin~Dela Rosa, Derek~Hao Hu, and Shih-Fu Chang.
\newblock Open-vocabulary object detection using captions.
\newblock In {\em Proceedings of the IEEE/CVF Conference on Computer Vision and
  Pattern Recognition (CVPR)}, pages 14393--14402, 2021.

\bibitem{zhang2021depth}
Zhijie Zhang, Yan Liu, Junjie Chen, Li Niu, and Liqing Zhang.
\newblock Depth privileged object detection in indoor scenes via deformation
  hallucination.
\newblock In {\em Proceedings of the AAAI Conference on Artificial
  Intelligence}, volume~35, pages 3456--3464, 2021.

\bibitem{zhou2022scale}
Feng Zhou, Yu-Kun Lai, Paul~L Rosin, Fengquan Zhang, and Yong Hu.
\newblock Scale-aware network with modality-awareness for rgb-d indoor semantic
  segmentation.
\newblock {\em Neurocomputing}, 492:464--473, 2022.

\bibitem{zhu2019pdnet}
Chunbiao Zhu, Xing Cai, Kan Huang, Thomas~H Li, and Ge Li.
\newblock Pdnet: Prior-model guided depth-enhanced network for salient object
  detection.
\newblock In {\em 2019 IEEE International Conference on Multimedia and Expo
  (ICME)}, pages 199--204. IEEE, 2019.

\end{thebibliography}
